\documentclass{article}

\usepackage{PRIMEarxiv}

\usepackage{threeparttable}
\usepackage{biblatex}
\usepackage{graphicx}
\usepackage{subfigure}
\usepackage[utf8]{inputenc} 
\usepackage[T1]{fontenc}    
\usepackage{hyperref}       
\usepackage{url}            
\usepackage{booktabs}       
\usepackage{amsfonts}       
\usepackage{nicefrac}       
\usepackage{microtype}      
\usepackage{xcolor}         
\usepackage{amsmath,amsfonts,bm,amsthm}
\usepackage{xfrac}

\newcommand{\x}{{\mathbf{x}}}
\newcommand{\y}{{\mathbf{y}}}
\newcommand{\z}{{\mathbf{z}}}
\newcommand{\w}{{\mathbf{w}}}
\newcommand{\p}{{\mathbf{p}}}
\newcommand{\ba}{{\mathbf{a}}}

\newcommand{\beps}{{\boldsymbol{\epsilon}}}

\newcommand{\mulambda}[1]{\mu_{\lambda, #1}}
\newcommand{\alambda}{a_\lambda}

\newcommand{\gauss}{{\boldsymbol{\xi}}}
\newcommand{\gz}{{\boldsymbol{\zeta}}}
\newcommand{\argmin}{\mathop{\mathrm{argmin}}}

\newcommand{\I}{{\mathbf{I}}}
\newcommand{\A}{{\mathbf{A}}}
\newcommand{\B}{{\mathbf{B}}}
\newcommand{\U}{{\mathbf{U}}}
\newcommand{\D}{{\mathbf{D}}}
\newcommand{\C}{{\mathbf{C}}}
\newcommand{\W}{{\mathbf{W}}}
\newcommand{\bXi}{\boldsymbol\Xi}

\newcommand{\bbeta}{\bm{\beta}}

\newcommand{\N}{{\mathbb{N}}}
\newcommand{\R}{{\mathbb{R}}}
\newcommand{\E}{{\mathbb{E}}}

\newcommand{\tr}{\mathrm{tr}}

\newcommand{\AGD}{\mathrm{AGD}}
\newcommand{\cO}{\mathcal{O}}

\newcommand{\effdim}{r}

\usepackage{algorithm}
\usepackage{algorithmic}

\theoremstyle{plain}
\newtheorem{theorem}{Theorem}[section]
\newtheorem{proposition}[theorem]{Proposition}
\newtheorem{lemma}[theorem]{Lemma}
\newtheorem{corollary}[theorem]{Corollary}
\newtheorem{definition}[theorem]{Definition}
\newtheorem{assumption}[theorem]{Assumption}
\newtheorem{remark}[theorem]{Remark}

\definecolor{myred}{HTML}{ae1908}
\newcommand{\myred}[1]{{\color{myred} #1}}

\title{CORE:~Common Random Reconstruction for \\ Distributed  Optimization with Provable\\
Low Communication Complexity
}

\author{
  Pengyun Yue \qquad Hanzhen Zhao \\
  Peking University\\
  \texttt{\{yuepy, hzzhao\}@pku.edu.cn} \\
    \And
  Cong Fang \qquad Di He\\
  Peking University\\
  \texttt{\{fangcong, dihe\}@pku.edu.cn} \\
   \And
  Liwei Wang \qquad Zhouchen Lin \qquad Song-Chun Zhu\\
  Peking University\\
  \texttt{\{wanglw, zlin, s.c.zhu\}@pku.edu.cn} \\
}

\begin{document}
\maketitle

\begin{abstract}
With distributed machine learning being a prominent technique for large-scale machine learning tasks, communication complexity has become a major bottleneck for speeding up training and scaling up machine numbers. In this paper, we propose a new technique named Common randOm REconstruction(CORE), which can be used to compress the information transmitted between machines in order to reduce communication complexity without other strict conditions. Especially, our technique CORE projects the vector-valued information to a low-dimensional one through common random vectors and reconstructs the information with the same random noises after communication. We apply CORE to two distributed tasks, respectively convex optimization on linear models and generic non-convex optimization, and design new distributed algorithms, which achieve provably lower communication complexities.  For example, we show for linear models CORE-based algorithm can encode the gradient vector to $\cO(1)$-bits (against $\cO(d)$), with the convergence rate not worse, preceding the existing results.
\end{abstract}


\section{Introduction}
Distributed machine learning and optimization have become the main technique for solving tasks with large model and data scales. In simple terms, the distributed optimization problem in machine learning can be regarded as minimizing an objective function $f$ defined as an average of individual functions that are respectively accessible by their corresponding local machines.  More specifically, we consider a constrained optimization problem
\begin{equation}\label{eq:problem}
  \begin{split}   \mathop{\mathrm{minimize}}_{\x\in \R^d} ~~&f(\x) \equiv \frac{1}{n}\sum_{i=1}^n f_i(\x_i)\\
\mathrm{s.t.}~~&\x_1 =\x_2=\cdots=\x_n. 
\end{split}  
\end{equation}
Here  $f_i$  represents the individual objective function at the local machine $i$ and the constraint in \eqref{eq:problem} guarantees different machines corporately finding the same minimizer of the global objective function $f$. 
Typical examples for $f_i$ include regression or classification over linear, graphic,  as well as    (deep) neural network models. In these cases,  $f_i$ shares the form as $f_i(\x) \equiv F(\x;\zeta_i)$, where $\zeta_i$ denotes the data stored in machine $i$ and $F$ represents the learning model. 

One dominating \textbf{bottleneck} for further improving the speed of distributed machine learning is the communication bandwidth. With the increase of machine numbers and parameter scale, time spent on communication can not be ignored and even becomes much longer than that on computation. Such a problem is much more salient when the bandwidth of computing cluster is restricted, such as mobile devices. Many researchers have noticed that reducing the dimensions of data transmitted between machines can effectively reduce the communication complexity, and proposed heuristics techniques, such as quantization \cite{seide20141} and sparsity \cite{Aji_2017},  to release the communication burden to some degree. Some more complete and theoretically guaranteed algorithms based on these techniques are proposed soon, but the convergence rate of these algorithms is always restricted. To the best of our knowledge, although some researches show how to improve existing compression techniques or propose several new ones, few results provide concrete and feasible compression techniques that can provably reduce communication costs and maintain algorithm accuracy under mild conditions. In this paper, we propose a new technique named Common randOm REconstruction (CORE) which presents a provable result on low communication complexity. CORE is a technique that can be used to transmit a sequence of vector-valued information that follows from well-known ideas from information theory and communication complexity theory, taking advantage of common random variables. At each round, the vector-valued information is projected to a low-dimensional vector using Gaussian random noises by the sender, and after communication reconstructed with the same noises by the receiver.  We show such a procedure generates an unbiased estimator of the original vector-valued information with a controlled variance. We apply CORE to two distributed tasks, namely convex optimization on linear models and generic non-convex optimization. 
Compared with some existing relevant researches, ours has certain advantages. First, we propose a concrete and feasible compression technique and algorithms instead of an abstract but potentially not implementable framework to reduce communication costs. Second, our algorithms achieve provably much lower communication costs compared with the existing algorithms under realizable conditions.

\subsection{Related Work}
\textbf{Gradient 
compression.}
Gradient compression is the main technique to reduce communication complexity during the process of training.  One of the representative achievements is the gradient quantization, for example, 1-bit SGD \cite{seide20141} and 1-bit Adam \cite{tang20211}, which heuristically compresses each component of the gradient into an integer that can be encoded in a few bits.  On this basis, TernGrad \cite{wen2017terngrad}, QSGD \cite{alistarh2017qsgd}, ECQ-SGD \cite{wu2018error}, ALQ \cite{faghri2020adaptive}, Natural Compression \cite{horvoth2022natural} and DIANA \cite{mishchenko2019distributed} further improved the gradient quantization by adding hyperparameters or combining with the adaptive technique to control the compression ratio. Another main technique is gradient sparsification, which transmits the main dimensions of the gradient instead of the whole. Top-K \cite{wangni2018gradient, shi2019distributed, jiang2018linear} was the main idea of gradient sparsification which chose the first $k$ larger dimensions of the gradient to transmit. Gradient Dropping \cite{Aji_2017}, DGC \cite{lin2017deep}, Atomo \cite{wang2018atomo} and IBCD \cite{mishchenko202099} obtained better results on this basis. In addition, some new techniques based on other ideas have also been developed and achieved good results. For example, PowerSGD \cite{vogels2019powersgd} proposed a new low-rank gradient compressor. SignSGD \cite{bernstein2018signsgd, safaryan2019stochastic} proposed a sign-based method with simple compression rules. A biased contractive compressor \cite{beznosikov2020biased}, a general class of unbiased quantization operators \cite{horvath2023stochastic} and three-point compressors (3PC) \cite{richtarik20223pc} were also proposed as innovative new achievements. However, the second moments of these estimations are often of order $d$, which implies a restriction of the total communication costs. 

\textbf{Random sketching.} 
Sketching \cite{gribonval2020sketching,woodruff2014sketching, ikonomovska2007survey} is a widely-used technique in machine learning, data mining, and optimization, whose core idea is to reduce the scale by a probabilistic data structure to approximate the data to reduce the computation costs. It is worth noticing that some researchers have started to use the sketching technique to reduce communication costs during the process of training. Specifically, \textcite{konevcny2016federated} proposed FedAvg to reduce the communication costs, which uses a random subset of the value of the full gradient to communicate. They call this method sketched update and integrate quantization (before random sketch) in experiments. \textcite{jiang2018sketchml} also proposed a quantization-based sketched gradient compression method, which divides the values of the gradient into four buckets bounded by quantiles and encodes them. However, there is a lack of theoretical guarantees in the convergence of these algorithms, and these methods are still based on random sampling and heuristic quantization encoding. Moreover, \textcite{ivkin2019communication} proposed SKETCHED-SGD, which uses Count Sketch \cite{charikar2004finding} to compress the gradient. They also presented a theoretical analysis of convergence, but compared with vanilla SGD, it requires an $\tilde\cO(\frac{1}{T}+\frac{d^2}{k^2T^2} + \frac{d^3}{k^3T^3})$ (where $d$ is the dimension of gradient and $k$ is a fixed parameter satisfying $k \leq d$) convergence rate. When $d$ is large, it is much worse than SGD. \textcite{rothchild2020fetchsgd} proposed FetchSGD which combines Count Sketch \cite{charikar2004finding} and Top-$k$ \cite{lin2017deep} for $k$-sparsification. It requires the same convergence rate as SGD for non-convex objective functions, but the communication cost is also dependent on $d$ at least. \textcite{hanzely2018sega} proved that when adding biased estimates on the basis of random matrix sketching, their algorithm achieves a faster convergence rate and can be accelerated. However, they did not come up with a specific sketching method. Moreover, \textcite{lee2019solving} and \textcite{pilanci2015newton} proposed some sketched Hessian-based second-order optimization algorithms. In this work, we mainly focus on gradient-based communication-efficient methods.

\textbf{Distributed optimization.}
Distributed machine learning and optimization have developed rapidly in recent years. In the early years, the main achievements were based on the existing optimization algorithms, such as SGD with mini-batch \cite{cotter2011better}, DSVRG \cite{lee2015distributed}, EXTRA \cite{shi2015extra} and MSDA \cite{scaman2017optimal}.  In recent years, it is worth noticing that some compressed distributed optimization methods have been proposed. 
Inspired by the results of coordinate gradient descent, \textcite{safaryan2021smoothness, hanzely2018sega} proposed a compressed gradient descent framework based on the idea of random projection, achieving an $\tilde\cO(\frac{\sum_{i=1}^d M_{ii}^{1/2}}{\mu^{1/2}})$ communication complexity as \textcite{allen2016even} for $\mathbf{M}$-Hessian dominated and $\mu$-strongly convex function where $M_{ii}$ is the entry in $i$-th row and $i$-th column of the matrix $\mathbf{M}$, but their achievements are lack of a concrete projection compression method.  Moreover, some compressed gradient descent algorithms based on compression techniques mentioned above were also proposed, such as DCGD \cite{khirirat2018distributed}, DIANA \cite{mishchenko2019distributed}, MARINA \cite{gorbunov2021marina}, DASHA \cite{tyurin2022dasha} and CANITA \cite{li2021canita}. \textcite{li2020acceleration} proposed an accelerated distributed algorithm which achieved $\tilde\cO(d + \frac{d L^{1/2}}{n\mu^{1/2}})$ considering the setting that the dimension $d$ is extremely larger than $n$, whose lower bound is $\tilde\cO(d + \frac{d^{1/2} L^{1/2}}{\mu^{1/2}})$. It is worth noticing that in practice $d$ is often extremely large.  So there is still a lack of a concrete compression technique and corresponding distributed algorithm that achieves low communication complexity when $d$ is large. And our work fills this gap. In addition, error feedback technique \cite{stich2019error, karimireddy2019error, tang2019doublesqueeze, gruntkowska2022ef21, richtarik2021ef21, fatkhullin2021ef21} was also widely used in compressed distributed optimization.  

\textbf{Federated learning.} Federated Learning is another machine learning setting concentrating on communication costs, where the goal is to train a high-quality centralized model while training data remains distributed over a large number of clients each with unreliable and relatively slow network connections. In federated learning communication bandwidth is a dominating bottleneck and how to design a communication-efficient algorithm has been a concern for researchers. In the early years, some federated learning algorithms have been proposed. These methods are often based on local gradient and heuristic gradient compressors to reduce one-step communication cost, such as FedAvg \cite{konevcny2016federated}, FetchSGD \cite{rothchild2020fetchsgd}, SKETCHED-SGD \cite{ivkin2019communication}, SCAFFOLD \cite{karimireddy2020scaffold} and FedLin \cite{mitra2021linear}. However, the approximation of local gradient often results in a loss of convergence rate. The total communication costs are worse than, or at best matching that of vanilla gradient descent. Recently, some new communication-efficient methods such as Scaffnew \cite{mishchenko2022proxskip} and GradSkip \cite{maranjyan2022gradskip} have been proposed to achieve the same communication rounds as the lower bound of smooth and strongly-convex objective functions $\cO(\sqrt{\kappa})$, but the total communication costs are still $\cO(d)$.

\textbf{Random communication complexity. }
In theoretical computer science, communication complexity studies the amount of communication needed to solve a problem when input data is distributed among several parties. Communication complexity was first proposed in \cite{andrew1979some}. \textcite{andrew1979some} also defined randomized protocol and randomized communication complexity. In a randomized protocol, parties are given a common random string as the input to a deterministic protocol. Random protocols can determine the answer in high probability with much less amount of information transmitted, so randomized communication complexity is much lower than deterministic communication complexity in expectation. Inspired by the advantage of randomized protocols over deterministic ones, we designed a random compression method for distributed optimization which is faster in expectation. \textcite{newman1991private} proved that any protocol using a common random string can be simulated by a private random string protocol, with an extra $\cO(\log n)$ bits.
\subsection{Contributions}\label{sec:contribution}
In this work, we propose the Common randOm REconstruction (CORE) technique and apply CORE to two distributed tasks, which present the advantages as below:

To the best of our knowledge, CORE is a  concrete and feasible compression method that can achieve a limited bounded variance of the estimate and probably reduce communication complexity when the eigenvalues of the Hessian matrices of $f$ drop very fast, which is currently in short.  We notice that in practice the situation that the eigenvalues of Hessian decrease fast has been considered for a long time, such as one defines the effective rank (e.g. \textcite{hsu2012random}) to represent the acting dimension of the data on linear models. Some recent empirical studies, for instance,  \textcite{sagun2016eigenvalues} carefully compute the eigenvalue of Hessian curves during training for (deep) neural networks. (See Figure \ref{fig4}: an example of eigenvalues for a real dataset and a neural network in Appendix \ref{sec: fig}). 

To characterize the strength of CORE in rigor,  we introduce the factor 
\begin{equation}
    r_{\alpha} = \sup_{\x\in \mathbb{R}^d} \sum_{i=1}^d \lambda_i^\alpha(\nabla^2 f(\x)),\qquad \alpha>0
\end{equation} as the effective dimension for distributed optimization, where $\lambda_i(\cdot)$ is the $i$-th singular value (also the eigenvalue when $\nabla^2 f(\x)$ is semi-definite in convex case). 
This is inspired by the recent work from Zeroth-order optimization \cite{yue2023zerothorder}, Langevin sampling \cite{freund2022convergence}, and distributed optimization \cite{hanzely2018sega}. We also introduce the Hessian domination assumption which is used in several researches \cite{hanzely2018sega, safaryan2021smoothness,yue2023zerothorder} for theoretical analysis. We propose that CORE can be applied to some gradient-descent-based algorithms. We examine linear models which are basic in machine learning and contain examples such as kernel regression and (deep) neural networks in the neural tangent kernel regime. By combining CORE with centralized gradient descent (CGD), we propose the CORE-Gradient Descent (CORE-GD) algorithm and prove that for the standard case where $f$ has $L$-Lipschitz gradients,  CORE-GD achieves $\cO\left(r_1(f)D^2\epsilon^{-1}\right)$ communication costs to obtain an $\epsilon$-optimal solution, where $D = \|\x^0 -\x^*\| $. Compared with CDG which achieves $\cO\left(dLD^2\epsilon^{-1}\right)$ communication costs, CORE-GD has a significant advantage since $r_1(f)$ is much smaller than $dL$ in most cases when eigenvalues decay fast. 
In Appendix \ref{sec: acc},  we also study accelerations of CORE-GD using the momentum technique, 
and propose a heavy-ball-based accelerated algorithm named CORE-Accelerated Gradient Descent (CORE-AGD).
We prove that on the linear regression model, CORE-AGD achieves the state-of-the-art
${\tilde\cO}\left(\frac{r_{1/2}(f)}{\mu^{1/2}}\right)$ communication costs which is lower than $\tilde\cO(d + \frac{d L^{1/2}}{n^{1/2}\mu^{1/2}})$ in \textcite{li2020acceleration} and ${\tilde\cO}\left(\frac{\sum_{i=1}^d M_{ii}^{1/2}}{\mu^{1/2}}\right)$ in  \textcite{hanzely2018sega}. More details and comparisons are shown in Table \ref{table:comparison}. Compared with the results in \textcite{hanzely2018sega} , our works present a concrete compression technique.
In Section \ref{sec:non-convex}, we then examine the efficiency of CORE in generic non-convex optimization when finding an $\epsilon$-approximated first-order stationary point. 
We further assume a Hessian-Lipschitz condition and show that  CORE-GD with carefully chosen stepsize 
can achieve lower communication costs which reduces upon the communication costs of CGD by a $\min\left\{dL/r_1(f), \epsilon^{-0.5}d^{1/4} \right\}$ factor.

In summary, the contribution of the paper is listed below:
\begin{itemize}\vspace{-0.1in}
    \item[(A)] We propose a new technique called CORE to efficiently transmit information between machines. To the best of our knowledge, CORE is the \textit{first} concrete and feasible compression technique that is  provably more efficient on communication when eigenvalues drop fast and can be applied to gradient-descent-based algorithms.  
    
    \item[(B)] We apply CORE to  convex optimization on linear models and generic non-convex optimization. We design new optimization algorithms and show 
 a \textit{remarkable reduction} of communication complexity under realizable conditions. Compared with the recent distributed optimization and federated learning algorithms, our CORE-GD and CORE-AGD achieve the lower bound of iteration rounds the \textit{state-of-the-art} total communication costs under the realizable condition. 
 
\end{itemize}
Finally, we propose a reduction framework that extends CORE to work on decentralized communication in Appendix \ref{sec:decentralized}.  We show the price is  only an additional $\tilde{\cO}(\sqrt{\gamma})$ factor, where $\gamma$ is the eigengap of the gossip matrix for the network topology. We also show that CORE is equipped with some privacy guarantee naturally for the use of random vectors, and prove our results in Appendix \ref{differentialprivacy}. We conduct empirical studies where we compare CORE with the basic frequently used quantization and sparsity techniques both on linear models and (deep) neural networks in Appendix \ref{sec:apex}. 

\begin{table}[t]
    \centering
    \caption{The performance of communication-efficient methods }
    \begin{threeparttable}

    \begin{tabular}{llcccc}
        
        \toprule
        & method & communication rounds  & compressor & floats sent per round & total communication costs  \\
        \midrule
        & CGD \cite{nesterov2003introductory} & $\tilde\cO(\frac{L}{\mu})$ & - & $\Theta(d)$ & $\tilde\cO(\frac{dL}{\mu})$\\
        & ACGD \cite{nesterov2003introductory} & $\tilde\cO(\frac{L^{1/2}}{\mu^{1/2}})$ & - & $\Theta(d)$ & $\tilde\cO(\frac{dL^{1/2}}{\mu^{1/2}})$\\
        & FedLin \cite{mitra2021linear} & $\tilde\cO(\frac{d^{3/2}L}{k^{3/2}\mu})$ & Top-K\tnote{1}  & $\Theta(k)$ & $\tilde\cO(\frac{d^{3/2}L}{k^{1/2}\mu})$\\ 
        & Scaffnew \cite{mishchenko2022proxskip} & $\tilde\cO(\frac{L^{1/2}}{\mu^{1/2}})$ & Skip \tnote{2} & $\Theta(d)$& $\tilde\cO(\frac{dL^{1/2}}{\mu^{1/2}})$ \\
        & GandSkip \cite{maranjyan2022gradskip} & $\tilde\cO(\frac{L^{1/2}}{\mu^{1/2}})$ & Skip \tnote{2} & $\Theta(d)$& $\tilde\cO(\frac{dL^{1/2}}{\mu^{1/2}})$ \\
        & DIANA \cite{mishchenko2019distributed} & $\tilde\cO(\frac{d}{K}+\frac{dL}{Kn\mu})$ \tnote{3}& Top-K \tnote{1}& $\Theta(K)$& $\tilde\cO(d+\frac{dL}{n\mu})$ \\
        & ADIANA \cite{li2020acceleration} & $\tilde\cO(\frac{d}{K}+\frac{dL^{1/2}}{Kn^{1/2}\mu^{1/2}})$ \tnote{3}& Top-K \tnote{1}& $\Theta(K)$& $\tilde\cO(d+\frac{dL^{1/2}}{n^{1/2}\mu^{1/2}})$ \tnote{4} \\
        & ASEGA \cite{hanzely2018sega} & $\tilde\cO(\frac{\sum_{i=1}^d A_{ii}^{1/2}}{\mu^{1/2}})$ & - & $\Theta(1)$ \tnote{5}& $\tilde\cO(\frac{\sum_{i=1}^d A_{ii}^{1/2}}{\mu^{1/2}})$\\
        & CORE-GD (this work) & $\tilde\cO(\frac{L}{\mu})$ & CORE & $\Theta(\frac{\tr(\A)}{L})$& $\tilde\cO(\frac{\tr(\A)}{\mu})$ \\
        & CORE-AGD (this work) & $\tilde\cO(\frac{L^{1/2}}{\mu^{1/2}})$ & CORE & $\Theta(\frac{\sum_{i=1}^d \lambda_{i}^{1/2}}{L^{1/2}})$& $\tilde\cO(\frac{\sum_{i=1}^d \lambda_{i}^{1/2}}{\mu^{1/2}})$ \\
        \bottomrule
    \end{tabular}
    \begin{tablenotes}    
        \footnotesize               
        \item[1] FedLin, DIANA and ADIANA only propose the algorithms using compressor but not propose concrete gradient compression technique. Correspondingly, they use Top-K as an example to analyse the communication rounds and costs.
        \item[2] Scaffnew and GandSkip use communication skipping instead of gradient compressor. Specifically, they only communicate every $\cO(\frac{L^{1/2}}{\mu^{1/2}})$ rounds and the total computation rounds are $\tilde\cO(\frac{L}{\mu})$. 
        \item[3] The communication rounds of DIANA are $\tilde\cO(\omega + \frac{\omega L}{n\mu})$ when $\omega \geq n$. And similarly, that of ADIANA is $\tilde\cO(\omega + \frac{\omega L^{1/2}}{n^{1/2}\mu^{1/2}})$ when $\omega \geq n$. Here $\omega$ is compression ratio. For example when using Top-K compressor it is $\frac{d}{K}$, which is much larger than $n$ when the dimension of data is extremely large. In this setting $n$ can be seen as $\cO(1)$.
        \item[4] The lower bound of the total communication costs of this method is $\tilde\cO(d+\frac{d^{1/2}L^{1/2}}{\mu^{1/2}})$, and the upper bound of CORE-AGD is $\tilde\cO(\frac{d^{1/2}\tr(\A)^{1/2}}{\mu^{1/2}})$ for Cauchy-Schwarz inequality. In most cases when $\tr(\A)$ is bounded and $d$ is much large, CORE-AGD is better.
        \item[5] This method is coordinate-descent-based. CORE-AGD is better than it. Letting $\A=\U^\top \Sigma \U$ where $\U=[u_{ij}]$ and $\Sigma = diag\{\lambda_i\}$, we have $A_{ii} = \sum_{j=1}^d \lambda_j u_{ji}^2  \geq (\sum_{j=1}^d \lambda_j^{1/2} u_{ji}^2)^2$ (because the Hessian matrix is positive definite and symmetric). Thus we have $\sum_{i=1}^d A_{ii}^{1/2} \geq  \sum_{i=1}^d \lambda_i^{1/2}$.
    \end{tablenotes} 

    \end{threeparttable}
    \label{table:comparison}
\end{table}

\textbf{Notation.} Throughout this paper, we use the convention $\mathcal{O}\left( \cdot \right)$,   $\Omega\left(\cdot\right)$, and $\Theta\left(\cdot\right)$ to denote the \textit{lower}, \textit{upper} and \textit{lower and upper} bound with a global constant, and use $\tilde{\mathcal{O}}(\cdot)$ to denote the lower bound that hides a poly-logarithmic factor of the parameters. 
Let $\R$ denote the set of real numbers, and $\R^d$ denote a $d$-dimensional Euclidean space. We use bold lowercase letters, like $\x$, to represent a vector, and bold capital letters, like $\A$, to represent a matrix. Specially, we use $\I_d$ to represent the identity matrix in $d$-dimensional Euclidean space, and omit the subscript when $d$ is clear from the context for simplicity. Let $\langle\cdot,\cdot\rangle$ denote the inner product of two vectors in the Euclidean space, $\|\x\|$ denote the Euclidean norm of a vector, and $\|\A\|$ denote the operator norm of a matrix. It is worth noticing that we use $\|\x\|_\A$ to denote the Mahalanobis (semi) norm  where $\A$ is a positive semi-definite matrix, which can be specifically defined as
\begin{equation}
    \|\x\|_\A = \sqrt{\x^\top\A\x}.
\end{equation}

For all the functions $f$ appearing in this paper, we simply assume that $f\in \mathcal C^2$, which means that $f$ is  second-order derivative. We use $\nabla f(\x)$ and $\nabla^2 f(\x)$ to denote the first-order and second-order derivative of $f$. Moreover, we always assume that the objective function $f$ satisfies some basic assumptions in (\ref{basicass}) and the minimizer of $f$ exists. We use  $\x^*$ to denote  the minimizer, i.e. $\x^*\overset{\triangle}=\argmin_\x f(\x)$ and   $f^*$ to denote its minimum value, i.e. $f^*\overset{\triangle}=\min_\x f(\x)$.

\section{Preliminary}
\label{basicass}
In this section, we formally present some definitions and assumptions to constrain the objective function and the optimization problem. 

\begin{assumption}[$L$-smoothness]
     We say a function $f$ is $L$-smooth (or has $L$-Lipschitz continuous gradients), if 
    \begin{equation}\notag
        \|\nabla f(\x)-\nabla f(\y)\| \le L\|\x-\y\|, \quad \forall \x,\y \in \R^d.
    \end{equation}
\end{assumption}
Consequently, for the function $f\in \mathcal C^2$, we have the following two inequalities based on the $L$-smoothness of $f$ (see \textcite[Chapter 1]{nesterov2003introductory}):
     $$   f(\y)\le f(\x) + \langle \nabla f(\x), \y-\x\rangle + \frac{L}{2}\|\x-\y\|^2, \quad \forall \x,\y \in \R^d.$$

\begin{assumption}[Convexity]
    We say a function $f$ is convex if
    \begin{equation}\notag
         f(\y)\ge f(\x) + \langle \nabla f(\x), \y-\x\rangle + \frac{\mu}{2}\|\x-\y\|^2,\quad \forall \x,\y \in \R^d,
         \label{equ:convex}
    \end{equation}
    where $\mu\ge 0$. Moreover,  if $\mu>0$,  $f$ is said to be $\mu$-strongly convex. 
\end{assumption}

\begin{assumption}[$H$-Hessian Lipschitz continuity]
     We say $f\in \mathcal C^2$ has $H$-Hessian Lipschitz   continuous Hessian matrices if 
    \begin{equation}\notag
        \|\nabla^2 f(\x)-\nabla^2 f(\y)\|\le H\|\x-\y\|, \quad \forall \x,\y \in \R^d.
    \end{equation}
\end{assumption}

Next we define some frequently-used criteria for an approximate solution.

For convex problems, we aim to find an $\epsilon$-approximate solution satisfying the definition below:
\begin{definition}[$\epsilon$-approximate solution]
    We say $\x$ is an $\epsilon$-approximate solution of $f $ if
    \begin{equation}\notag
    f(\x)-f^*\le\epsilon.
    \end{equation}
\end{definition}

For non-convex problems, finding an $\epsilon$-approximate solution in general is NP-hard \cite{murty1985some}. Instead we consider finding an $\epsilon$-approximate first-order stationary point satisfying the definition below:
\begin{definition}[$\epsilon$-stationary point]
    We say $\x$ is an $\epsilon$-appriximate first-order stationary point of $f$ if 
    \begin{equation}\notag
    \|\nabla f(\x)\|\le\epsilon.
    \end{equation}
\end{definition}

\section{Common Random Reconstruction: Core Idea}\label{sec:core idea}
In this section, we present in detail the underlying idea of our Common RandOm REconstruction (CORE) technique behind the algorithm design. We can see such a technique reduces the quantities of data transmitted during communication to a great extent, which significantly reduces the communication complexity. It is of great importance in distributed optimization tasks. 


In most distributed machine learning tasks, information is transferred from one machine to another one in vector form, i.e. the gradient of the objective function. Suppose the dimension of the information is $d$. When a machine transmits a $d$-dimensional vector to another machine, we think the communication cost is $d$. However, in most applications, the dimension $d$ is very large. As a result, it is very expensive to send the whole vector. Inspired by the theory of communication complexity \cite{andrew1979some}, we propose a \textbf{feasible technique which realizes the dimension reduction by randomization}. Specifically, we suppose that all the machines have a common random number generator, which generates a fresh random Gaussian vector $\gauss\sim N(0, \I_d)$ at each transmission. We denote the information we want to transmit by $\ba\in \R^d$. Instead of sending the $d$-dimension vector $\ba$, we send a scalar $\langle \ba, \gauss\rangle$ which is the inner product of $\ba$ and the common random Gaussian vector $\gauss$. Then the receiver reconstructs $\ba$ by multiplying $\gauss$ with the scalar. 

To ensure the training accuracy and convergence rate, we can take $m$ fresh random Gaussian vectors for dimension reduction, where $m$ is the one-round communication budget.  Specifically, We send $m$ scalars which are the inner products of $\ba$ with $m$ random Gaussian vectors, and reconstruct $\tilde\ba$ by  averaging over the reconstructions using all $m$ random Gaussian vectors. We call this compression and reconstruction scheme Common Random Reconstruction (CORE), and describe it in  Algorithm \ref{alg:communication}.
\begin{algorithm}[t]
\caption{CORE: Common Random Reconstruction}\label{alg:communication}
\begin{algorithmic}
\REQUIRE{An vector $\ba$, machines $M_1$ and $M_2$, one-round communication budget $m$, a common random number generator}
\WHILE{ $M_1$ want to send $\ba$ to $M_2$}
\STATE{Generate  fresh i.i.d. random Gaussian vectors $\gauss_1,\cdots,\gauss_m\sim N(0,\I_d)$ with the common random number generator}
\STATE{$M_1$ sends $\{p_i\}_{i=1}^m$ to $M_2$ with $p_i=\langle\ba, \gauss_i\rangle$ }
\STATE{$M_2$ reconstructs $\ba$ by $\tilde \ba = \frac{1}{m}\sum_{i=1}^m p_i\cdot \gauss_i$}
\ENDWHILE
\end{algorithmic}
\end{algorithm}
In Algorithm \ref{alg:communication}, the estimation of $\ba$ admits:
\begin{equation}
    \tilde \ba = \frac{1}{m} \sum_{i=1}^m \langle \ba, \gauss_i\rangle \cdot\gauss_i.
    \label{equ:reconstruction}
\end{equation}
It is easy to spot that if we let the one-round communication budget $m$, the communication cost of each round is $m$. So the next important question is whether this technique can guarantee the accuracy of the results. In Lemma \ref{lem:unbiased} and Lemma \ref{lem:descent}, we show that $\tilde \ba$ is an unbiased estimator, and the variance of $\tilde\ba$ can be bounded under arbitrary matrix norms, respectively.


\begin{lemma}
$\tilde\ba$ is an unbiased estimator of $\ba$, 
\begin{equation}
    \E_{\gauss_1,\cdots\gauss_m} \tilde\ba = \ba.
\end{equation}
\label{lem:unbiased}
\end{lemma}

\begin{lemma}
The variance of $\tilde\ba$ under norm $\|\cdot\|_\A$, where $\A$ is a given positive semi-definite symmetric matrix, can be bounded by $\frac{3\tr(\A)}{m}\|\ba\|^2 - \frac{1}{m}\|\ba\|_\A^2$, 
\begin{equation}
    \E_{\gauss_1,\cdots,\gauss_m} \|\tilde \ba - \ba \|_\A^2 \le \frac{3\tr(\A)}{m}\|\ba\|^2 - \frac{1}{m}\|\ba\|_\A^2.
\end{equation}
\label{lem:descent}
\end{lemma}

\begin{remark}
Lemmas \ref{lem:unbiased} and \ref{lem:descent} bound the first and second moments of $\tilde \ba$, which provide us theoretical guarantee of the convergence accuracy if we replace $\ba$ by $\tilde\ba$ in certain algorithms. First, it is obvious that $\tilde \ba$ has a \textbf{sub-exponential tail} distribution given $\ba$, so we can provide high probability results using concentration inequalities. Second, the variance of $\tilde\ba$ is \textbf{lower bounded} when $\tr(\A)$ is smaller, ensuring the convergence accuracy of our technique with a lower communication cost.  
\end{remark}
In most cases when eigenvalues decrease fast which implies that $\tr(\A)$ is not large, our technique can achieve significant improvement. In fact, the CORE technique can be applied to a variety of distributed optimization tasks under different settings, for example, gradient-based algorithms or proximal algorithms, centralized or decentralized distributed optimization. In this paper, we mainly research the gradient-based distributed optimization algorithms on the centralized distributed optimization, by replacing the vector transmitted from the accurate coordinates of a gradient vector $\ba$ to the reconstruction by our CORE method, $\tilde \ba$ to reduce the communication cost in each round. 

\section{CORE on Linear Models}\label{sec: linear model}
In this section, we study the behavior of CORE on linear models. More specifically, to illustrate the CORE technique, we present some simple cases which are representative and contain the linear model,  perhaps the most important application of convex optimization in machine learning. For more involved cases, the analysis can be extended. For example, it is applied to general convex  optimization under an additional Hessian-smoothness condition using  methods like cubic regularization \cite{nesterov_cubic_2006} 
or our analysis in general non-convex optimization in Section \ref{sec:non-convex}. 



We start with the general components of CORE. Suppose we have $n$ machines. 
Based on the analysis of our core idea, we use Algorithm \ref{alg:communication} to compress and reconstruct the gradient vector as below, 
\begin{equation}
    \tilde  \nabla_m f(\x) = \frac{1}{nm} \sum_{i=1}^n \sum_{j=1}^m 
    \langle\nabla f_i(\x), \gauss_j \rangle\cdot\gauss_j.
    \label{equ:tildefdef}
\end{equation}
Then from Lemma \ref{lem:unbiased} and Lemma \ref{lem:descent},
$\tilde  \nabla_m f(\x)$ is an unbiased stochastic estimation of $\nabla f(\x)$ with a controlled variance. This implies that one can design a variety of optimization algorithms using the stochastic oracle $\tilde  \nabla_m f(\x)$. These algorithms can be efficiently implemented by CORE.  We introduce two typical  algorithms which are designed based on the GD and AGD respectively as follows.



Now we introduce the  CORE-GD algorithm, where at each gradient descent step,  the gradient $\nabla f(\x)$ is replaced by estimator $\tilde\nabla f(\x)$ using CORE.
The whole algorithm is presented in Algorithm \ref{alg:CORE-GD-minibatch}, where we let $m$ be the communication budget for a communication round. 

\begin{algorithm*}[t]
\caption{CORE-GD with per-round communication budget $m$}\label{alg:CORE-GD-minibatch}
\begin{algorithmic}
\REQUIRE{$n$ machines, a central machine, a common random number generator, $m\le \frac{\tr(\A)}{L}$, $\x^0$, $k=0$,  step-size $h_k = \frac{m}{4\tr(\A)}$}
\WHILE{$k<N$}
    \STATE{Generate fresh i.i.d. $m$ Gaussian vectors $\gauss_1,\cdots,\gauss_m$ with the common random number generator}
    \STATE{Machine $i$ sends $p_{ij}=\langle\nabla f_i(\x^k),\gauss_j \rangle$ to the central machine} 
    \STATE{The central machine sends $\sum_{i=1}^n p_{ij}$ back to every machine}
    \STATE{Machines reconstruct $\tilde \nabla_m f(\x^k)$ by $\tilde \nabla_m f(\x^k) = \frac{1}{m}\sum_{i=1}^n\sum_{j=1}^m p_{ij}\gauss_j$}
    \STATE{Machines update $\x^k$ by $\x^{k+1} = \x^k - h_k\tilde\nabla_m f(\x^k)$}
    \STATE{$k\gets k+1$}
\ENDWHILE
\end{algorithmic}
\end{algorithm*}

To show the strength of CORE,  we consider the objective function satisfying $\A$-Hessian domination condition defined as follows:
\begin{definition}[$\A$-Hessian domination]
     $f$ is said to be $\A$-Hessian dominated if there exists $\A$ such that
    \begin{equation}
        \nabla^2 f(\x) \preceq \A
    \end{equation}
    for every $\x\in\R^d$.
\label{AH}
\end{definition}
The idea is to characterize the complexity of our algorithm in terms of  $\tr(\A)$.  We note that when $f$ is $L$-smooth, a loose bound for $\A$ is $\A \preceq LI$. The fact implies that $\tr(\A)$ will reach  $dL$ in the worst case, whereas, $\tr(\A)$ can be much smaller than $dL$ in most cases. We will show that the linear models are $\A$-Hessian dominated. Moreover, when the data is normalized to a constant level, $\tr(\A)$ is much smaller and dimension-free. This result suggests only transmitting $\cO(1)$-bits information using CORE without lowering the convergence rate in expectation under suitable conditions.
We shall  mention that a similar idea of Hessian domination is also considered by \textcite{freund2022convergence} in the Langevin sampling algorithm, who instead proposes a squared Hessian domination condition.

We first consider the $\mu$-strongly convex case.  Theorem \ref{thm:CORE-GD-sconvex} below provides a linear convergence results for Algorithm~\ref{alg:CORE-GD-minibatch}.

\begin{theorem}
Suppose $f$ is $\mu$-strongly convex, $L$-smooth, and $\A$-Hessian dominated . Let $h_k = \frac{m}{4\tr(\A)}$. Then, under the hyper-parameter setting in Algorithm \ref{alg:CORE-GD-minibatch}, $\{\x^k\}_{k\in \N}$ satisfy for all $k\geq0$
\begin{equation}
    \E f(\x^{k+1})-f^* \le \left(1-\frac{3m\mu}{16\tr(\A)}\right)\left(f(\x^k)-f^*\right). 
\end{equation}
\label{thm:CORE-GD-sconvex}
\end{theorem}
\begin{remark}
    According to Theorem \ref{thm:CORE-GD-sconvex}, our total communication costs are $\cO\left(\frac{\tr(\A)}{\mu} \log \frac{1}{\epsilon} \right)$ in expectation. As we have mentioned, high probability results can also be obtained with additional logarithmic factors, which we simply omit here.
\end{remark}
 
\begin{remark}
    We compare CORE-GD with the vanilla CGD algorithm which has  total communication costs $\cO\left(\frac{dL}{\mu}\log\frac{1}{\epsilon}\right)$. CORE-GD achieves provably lower communication costs since we always have $\tr(\A)\leq dL$ when ignoring constants. CORE-GD is also better than DIANA \cite{mishchenko2019distributed} whose total communication cost is $\cO(d+\frac{dL}{n\mu})$ when $d$ is extremely larger than $n$. Moreover, Theorem \ref{thm:CORE-GD-sconvex}  requires the communication budget $m\leq \frac{\tr(\A)}{L}$. When $m =\Theta\left(\frac{\tr(\A)}{L}\right)$,  CORE-GD achieves the same number of communication rounds (convergence rate) as those of CGD when ignoring  constants. And it is clear that more communication budget cannot accelerate the convergence rate.
\end{remark}

Next we present realizable conditions for  linear models that ensure $\tr(\A)$ to be small.  We consider the objective admits the so-called ridge-separable form \cite{freund2022convergence}:
\begin{align}\label{eq: ridgeseparable}
    f(\x) \equiv \frac{1}{N}\sum_{i=1}^N \sigma_i(\bbeta_i^\top\x) + \frac{\alpha}{2}\|\x \|^2.
\end{align}
Here, we simply consider the $\ell_2$ norm regularizer. It is possible to generalize our results using proximal algorithms for other regularizers. In \eqref{eq: ridgeseparable}, $\bbeta_i$ is associated with the data, and $\sigma_i$ is associated with the loss function. We make the following assumptions:
\begin{assumption}\label{asp: activation}
The function $\sigma_i\in\mathcal{C}^2$ has a bounded second derivative, i.e. $\sigma_i''\leq L_0$ for all 
$i\in[n]$.
\end{assumption}
\begin{assumption}\label{asp: data}
For all $i\in[N]$, then norm of $\bbeta_i$ is bounded by $R$, i.e. $\|\bbeta_i\|^2\leq R$.
\end{assumption}
Note that  Assumption \ref{asp: data} can be realized  by  normalizing the data and Assumption \ref{asp: activation}  only requires the loss to have a bounded second derivative. We show $\tr(\A)$ is small as follows:
\begin{lemma}
For the objective function in form of  \eqref{eq: ridgeseparable}, under Assumptions \ref{asp: activation} and \ref{asp: data}, then $f$ is $\A$-Hessian dominated and $\A$ satisfies
\begin{equation} \label{eq: rshessianbound}
    \tr(\A) \leq d\alpha+ L_0R.
\end{equation}
\label{lemma45}
\end{lemma}
With Lemma \ref{lemma45}, we show CORE-GD ensures much low communication costs for linear models under suitable conditions.
\begin{corollary}
For the objective function in form of  \eqref{eq: ridgeseparable}, under Assumptions \ref{asp: activation} and \ref{asp: data}, with $\tr (\A)$ defined in \eqref{eq: rshessianbound},  
the total communication costs of CORE-GD are $\cO\left(\left(d+\frac{L_0R}{\alpha}\right)\log\frac{1}{\epsilon} \right)$. 
\label{cor:linear-model}
\end{corollary}
\begin{remark}
    From Corollary \ref{cor:linear-model}, treated $R$ and $L_0$ as constants, the total communication costs of CORE-GD  are $\tilde{\cO}(d+\alpha^{-1})$, whereas the vanilla CGD requires $\tilde{\cO}(d\alpha^{-1})$ communication costs. Here $\alpha^{-1}$ can be considered as the condition number of the objective since $L$ can be $\Theta(1)$.  CORE-GD greatly reduces the communication costs by the factor of $\min(d,\alpha^{-1})$. 
\end{remark}

\begin{algorithm*}[t]
\caption{CORE-GD in Non-convex Optimization}\label{alg:CORE-GDAS}
\begin{algorithmic}
\REQUIRE{$n$ machines, a central machine, a common random number generator, $m\le \frac{r_1(f)}{L}$, $\x^0$, $k=0$, (For Option I, $m>\log\left(\frac{N}{\delta}\right)$)}
\STATE{Assume that $f(\x^0)-f^*\le \Delta$}
\WHILE{$k<N$}
    \STATE{Generate fresh i.i.d. $m$ Gaussian vectors $\gauss_1,\cdots,\gauss_m$ with the common random number generator}
    \STATE{Machine $i$ sends $p_{ij}=\langle\nabla f_i(\x^k),\gauss_j \rangle$ to the central machine} 
    \STATE{The central machine sends $\sum_{i=1}^n p_{ij}$ back to every machine}
    \STATE{Machines reconstruct $\tilde \nabla_m f(\x^k)$ by $\tilde \nabla_m f(\x^k) = \frac{1}{m}\sum_{i=1}^n\sum_{j=1}^m p_{ij}\gauss_j$}
    \STATE{Let $p = \frac{1}{m}\sum_{i=1}^m \left(\sum_{j=1}^n p_{ij}\right)$}
    \STATE{ $h_k=\begin{cases}\min \{\frac{m}{16\effdim_1(f)}, \frac{1}{1600}H^{-1/2} p^{-1/2} d^{-3/4}m^{3/4}\},&\text{Option I}\\
    \min\{ \frac{m}{16\effdim_1(f)}, \frac{1}{1600}H^{-1/2} (L\Delta)^{-1/4} d^{-3/4}m^{3/4}\},&\text{Option II}  \end{cases}$}

    \STATE{$\tilde \x^{k+1} = \x^k - h_k\tilde\nabla_m f(\x^k)$}
    \STATE{$\x^{k+1}\gets \argmin_{\x\in \{\x_k,\tilde\x_{k+1}\}} f(\x)$}
    \STATE{$k\gets k+1$}
\ENDWHILE
\end{algorithmic}
\end{algorithm*}

We also consider the acceleration of our algorithm. Specifically, we consider heavy-ball \cite{polyak1964some} acceleration for CORE-GD for quadratic objective functions in Appendix \ref{sec: acc}. From Theorem \ref{thm:CORE-AGD},
the total communication costs to find an $\epsilon$-approximate solution in linear regression model for CORE-AGD are $\tilde{\cO}\left(\frac{\sum_{i=1}^d \lambda_i^{1/2}}{\mu^{1/2}}\right)$ , which is better than $\tilde\cO(d + \frac{d L^{1/2}}{\mu^{1/2}})$ because $\frac{\sum_{i=1}^d \lambda_i^{1/2}}{\mu^{1/2}} \le \frac{d^{1/2} \tr(\A)}{\mu^{1/2}}$. When $d$ is large and the trace of Hessian is bounded, this result is better than $\tilde\cO(d + \frac{d L^{1/2}}{\mu^{1/2}})$. And CORE-AGD is better than $\tilde\cO(\frac{\sum_{i=1}^d A_{ii}^{1/2}}{\mu^{1/2}})$ because $\sum_{i=1}^d \lambda_i^{1/2} \leq \sum_{i=1}^d A_{ii}^{1/2}$ based on the Cauchy-Schwarz inequality when $\A$ is semi-definite. Moreover, when $m = \Theta\left(\frac{\sum_{i=1}^d \lambda_i^{1/2}}{L^{1/2}}\right)$, CORE-AGD achieves the same number of communication rounds as those of Centralized AGD with ignoring logarithmic factors.

\newpage
\section{CORE-GD for Non-convex Optimization}\label{sec:non-convex}
In this section, we research CORE-GD on solving a general non-convex problem. To explore the information on Hessian matrices, we further assume that $f$ has $H$-Lipschitz continuous Hessian matrices. Moreover, we introduce  $\effdim_1(f)$ as a uniform upper bound of the trace of Hessian matrices, i.e.  $r_{1}(f) = \sup_{\x\in \mathbb{R}^d} \sum_{i=1}^d \lambda_i(\nabla^2 f(\x))$. $\effdim_1(f)$ is often much smaller than $dL$. (see Figure \ref{fig4} taken from \textcite{sagun2016eigenvalues} and empirical results in related papers, e.g. \textcite{sagun2017empirical, ghorbani2019investigation,brock2018large}.)  We will characterize the complexities of our algorithm in terms of  $\effdim_1(f)$. 

To be specific, we show that except linear models, more learning models have a limited $r_1(f)$. For example, we consider the two-layer neural network model as below. 

\begin{proposition}
Define $f(\W,\mathbf{w}) = \mathbf{w}^\top\sigma(\W\x)$, where $\sigma$ is the activation function. When $\| \x \| \leq a_1$, $\| \mathbf{w} \| \leq a_2$ and $\sigma^{''}(x) \leq \alpha$, we have $\tr(\nabla^2 f(\W, \mathbf{w}) \leq \alpha a_1 a_2$.
\end{proposition}

Moreover, we notice that for many parameterized models, $r_1(f)$ is limited at least when the parameter is close to its optimal solution. The reason is that under weak regular conditions, the fisher information $\mathcal{I}(\theta) = -\E\left[\frac{\partial^2}{\partial \theta^2}\log f(\mathbf{X};\theta) | \theta\right] = \E\left[\left(\frac{\partial}{\partial \theta}\log f(\mathbf{X}; \theta) \right)^2 | \theta \right]$. So when $\frac{\partial}{\partial \theta}\log f(\mathbf{X}; \theta) $ is limited, $r_1(f)$ is also limited, ensuring our results a wide range of applications.

 We consider the  CORE-Gradient Descent algorithm with some adaptations. The algorithm is shown in Algorithm \ref{alg:CORE-GDAS}. These adaptations are to guarantee low communication costs in theory. Specifically, we take a careful choice of step size. Moreover, we add one more comparison step, i.e. $\x^{k+1}\gets \argmin_{\x\in \{\x_k,\tilde\x_{k+1}\}} f(\x)$. The  step requires only one more round of  communication with  $\cO(1)$ communication costs.


Now we present the convergence rate and communication costs of CORE-GD in the non-convex setting. 

\begin{theorem}
    Assume that $f(\x)$ is $L$-smooth and has $H$-Lipschitz continuous Hessian matrix. With the assumption of $\tr(\nabla^2 f(\x))\le \effdim_1$ for any $\x\in\R^d$ and $f(\x^0)-f^*\le\Delta$. Then, under the hyper-parameter setting in Algorithm \ref{alg:CORE-GDAS}, the following result in expectation 
    \begin{equation}
        \E f(\x^k) \le f(\x^0) - \sum_{i=1}^k \E \left[ \frac{h_i}{2}\|\nabla f(\x^i)\|^2\right]
    \end{equation}
    holds for option II, and holds with probability $1-\delta$ for option I.
    \label{thm:CORE-GDAS}
\end{theorem}

\begin{remark}
With Theorem \ref{alg:CORE-GDAS}, we give the convergence rate and total communication costs of CORE-GD.
\begin{itemize}
        \item For Option I, CORE-GD needs $\cO\left(\max\left\{\frac{\Delta\effdim_1(f)}{m\epsilon^2}, \frac{\Delta H^{1/2}d^{3/4}}{m^{3/4}\epsilon^{3/2}} \right\}\right)$ rounds to find an $\epsilon$-stationary point with probability $1-\delta$. The total communication costs of CORE-GD are
        \begin{equation}\cO\left(\max\left\{\frac{\Delta\effdim_1(f)}{\epsilon^2}, \frac{\Delta H^{1/2}d^{3/4}m^{1/4}}{\epsilon^{3/2}} \right\}\right).\notag\end{equation}
        \item For Option II, CORE-GD needs $\cO\left(\max\left\{\frac{\Delta\effdim_1(f)}{m\epsilon^2}, \frac{\Delta^{5/4}L^{1/4} H^{1/2}d^{3/4}}{m^{3/4}\epsilon^{2}} \right\}\right)$ rounds to find an $\epsilon$-stationary point in high probability. The total communication costs of CORE-GD are
        \begin{equation}\cO\left(\max\left\{\frac{\Delta\effdim_1(f)}{\epsilon^2}, \frac{\Delta^{5/4}L^{1/4} H^{1/2}d^{3/4}m^{1/4}}{\epsilon^{2}} \right\}\right).\notag
        \end{equation}
    \end{itemize}
\end{remark}

\begin{remark}

Let us compare CORE-GD with Option I with vanilla CGD.
The communication costs of CGD to find an $\epsilon$-stationary point  is $\tilde{\cO}\left(dL\Delta\epsilon^{-2} \right)$. Treated $L$, $H$, $\Delta$ as constants,  when the per-round communication budget $m =\Theta\left(\frac{\tr(r_1(f))}{L}\right)$,  CORE-GD achieves the same number of communication rounds (convergence rate) as those of CGD,
CORE-GD with Option I reduces the communication costs by  a factor of $\min( dL/r_1,\epsilon^{-0.5}d^{1/4})$ when ignoring logarithmic factors. 
\end{remark} 

\section{Conclusion}\label{sec:conclusion}
In this paper, we propose the CORE technique to transmit information in distributed optimization which can dramatically reduce communication costs. We propose our CORE technique based on the common random variables, which provably reduce the quantities of information transmitted, and apply CORE to two distributed tasks. We prove that our CORE-based algorithms achieve lower communication costs. And by choosing the proper communication budget $m$, our algorithms can achieve the same number of communication rounds as those of uncompressed algorithms. In a word,  CORE provides new insights and opens the door for designing provably better compression methods in distributed optimization.

\newpage
\printbibliography

@article{yue2023zerothorder,
      title={Zeroth-order Optimization with Weak Dimension Dependency}, 
      author={Pengyun Yue and Long Yang and Cong Fang and Zhouchen Lin},
      year={2023},
      eprint={2307.05753},
      archivePrefix={arXiv},
      primaryClass={math.OC}
}

@article{pilanci2015newton,
      title={Newton Sketch: A Linear-time Optimization Algorithm with Linear-Quadratic Convergence}, 
      author={Mert Pilanci and Martin J. Wainwright and A and A},
      year={2015},
      eprint={1505.02250},
      archivePrefix={arXiv},
      primaryClass={math.OC}
}

@article{lee2019solving,
  title={Solving empirical risk minimization in the current matrix multiplication time},
  author={Lee, Yin Tat and Song, Zhao and Zhang, Qiuyi and A},
  journal={Conference on Learning Theory},
  pages={2140--2157},
  year={2019},
  organization={PMLR}
}

@article{maranjyan2022gradskip,
  title={Gradskip: Communication-accelerated local gradient methods with better computational complexity},
  author={Maranjyan, Artavazd and Safaryan, Mher and Richt{\'a}rik, Peter},
  journal={arXiv preprint arXiv:2210.16402},
  year={2022}
}

@article{mishchenko2022proxskip,
  title={Proxskip: Yes! local gradient steps provably lead to communication acceleration! finally!},
  author={Mishchenko, Konstantin and Malinovsky, Grigory and Stich, Sebastian and Richt{\'a}rik, Peter},
  journal={International Conference on Machine Learning},
  pages={15750--15769},
  year={2022},
  organization={PMLR}
}

@article{mitra2021linear,
  title={Linear convergence in federated learning: Tackling client heterogeneity and sparse gradients},
  author={Mitra, Aritra and Jaafar, Rayana and Pappas, George J and Hassani, Hamed},
  journal={Advances in Neural Information Processing Systems},
  volume={34},
  pages={14606--14619},
  year={2021}
}

@article{karimireddy2020scaffold,
  title={Scaffold: Stochastic controlled averaging for federated learning},
  author={Karimireddy, Sai Praneeth and Kale, Satyen and Mohri, Mehryar and Reddi, Sashank and Stich, Sebastian and Suresh, Ananda Theertha},
  journal={International Conference on Machine Learning},
  pages={5132--5143},
  year={2020},
  organization={PMLR}
}

@article{woodruff2014sketching,
  title={Sketching as a tool for numerical linear algebra},
  author={Woodruff, David P and others},
  journal={Foundations and Trends{\textregistered} in Theoretical Computer Science},
  volume={10},
  number={1--2},
  pages={1--157},
  year={2014},
  publisher={Now Publishers, Inc.}
}

@article{gribonval2020sketching,
  title={Sketching datasets for large-scale learning (long version)},
  author={Gribonval, R{\'e}mi and Chatalic, Antoine and Keriven, Nicolas and Schellekens, Vincent and Jacques, Laurent and Schniter, Philip},
  journal={arXiv preprint arXiv:2008.01839},
  year={2020}
}

@inproceedings{ikonomovska2007survey,
  title={A survey of stream data mining},
  author={Ikonomovska, Elena and Loshkovska, Suzana and Gjorgjevikj, Dejan},
  year={2007}
}

@article{konevcny2016federated,
  title={Federated learning: Strategies for improving communication efficiency},
  author={Kone{\v{c}}n{\`y}, Jakub and McMahan, H Brendan and Yu, Felix X and Richt{\'a}rik, Peter and Suresh, Ananda Theertha and Bacon, Dave},
  journal={arXiv preprint arXiv:1610.05492},
  year={2016}
}

@inproceedings{jiang2018sketchml,
  title={Sketchml: Accelerating distributed machine learning with data sketches},
  author={Jiang, Jiawei and Fu, Fangcheng and Yang, Tong and Cui, Bin},
  booktitle={Proceedings of the 2018 International Conference on Management of Data},
  pages={1269--1284},
  year={2018}
}

@article{ivkin2019communication,
  title={Communication-efficient distributed SGD with sketching},
  author={Ivkin, Nikita and Rothchild, Daniel and Ullah, Enayat and Stoica, Ion and Arora, Raman and others},
  journal={Advances in Neural Information Processing Systems},
  volume={32},
  year={2019}
}

@article{rothchild2020fetchsgd,
  title={Fetchsgd: Communication-efficient federated learning with sketching},
  author={Rothchild, Daniel and Panda, Ashwinee and Ullah, Enayat and Ivkin, Nikita and Stoica, Ion and Braverman, Vladimir and Gonzalez, Joseph and Arora, Raman},
  journal={International Conference on Machine Learning},
  pages={8253--8265},
  year={2020},
  organization={PMLR}
}

@article{charikar2004finding,
  title={Finding frequent items in data streams},
  author={Charikar, Moses and Chen, Kevin and Farach-Colton, Martin},
  journal={Theoretical Computer Science},
  volume={312},
  number={1},
  pages={3--15},
  year={2004},
  publisher={Elsevier}
}

@article{vargaftik2021drive,
  title={Drive: One-bit distributed mean estimation},
  author={Vargaftik, Shay and Ben-Basat, Ran and Portnoy, Amit and Mendelson, Gal and Ben-Itzhak, Yaniv and Mitzenmacher, Michael},
  journal={Advances in Neural Information Processing Systems},
  volume={34},
  pages={362--377},
  year={2021}
}

@article{stich2019error,
  title={The error-feedback framework: Better rates for SGD with delayed gradients and compressed communication},
  author={Stich, Sebastian U and Karimireddy, Sai Praneeth},
  journal={arXiv preprint arXiv:1909.05350},
  year={2019}
}

@inproceedings{tang2019doublesqueeze,
  title={Doublesqueeze: Parallel stochastic gradient descent with double-pass error-compensated compression},
  author={Tang, Hanlin and Yu, Chen and Lian, Xiangru and Zhang, Tong and Liu, Ji},
  booktitle={International Conference on Machine Learning},
  pages={6155--6165},
  year={2019},
  organization={PMLR}
}

@inproceedings{karimireddy2019error,
  title={Error feedback fixes signsgd and other gradient compression schemes},
  author={Karimireddy, Sai Praneeth and Rebjock, Quentin and Stich, Sebastian and Jaggi, Martin},
  booktitle={International Conference on Machine Learning},
  pages={3252--3261},
  year={2019},
  organization={PMLR}
}

@article{gruntkowska2022ef21,
  title={EF21-P and Friends: Improved Theoretical Communication Complexity for Distributed Optimization with Bidirectional Compression},
  author={Gruntkowska, Kaja and Tyurin, Alexander and Richt{\'a}rik, Peter},
  journal={arXiv preprint arXiv:2209.15218},
  year={2022}
}

@article{richtarik2021ef21,
  title={EF21: A new, simpler, theoretically better, and practically faster error feedback},
  author={Richt{\'a}rik, Peter and Sokolov, Igor and Fatkhullin, Ilyas},
  journal={Advances in Neural Information Processing Systems},
  volume={34},
  pages={4384--4396},
  year={2021}
}

@article{fatkhullin2021ef21,
  title={EF21 with bells \& whistles: Practical algorithmic extensions of modern error feedback},
  author={Fatkhullin, Ilyas and Sokolov, Igor and Gorbunov, Eduard and Li, Zhize and Richt{\'a}rik, Peter},
  journal={arXiv preprint arXiv:2110.03294},
  year={2021}
}

@article{safaryan2021smoothness,
  title={Smoothness matrices beat smoothness constants: Better communication compression techniques for distributed optimization},
  author={Safaryan, Mher and Hanzely, Filip and Richt{\'a}rik, Peter and A},
  journal={Advances in Neural Information Processing Systems},
  volume={34},
  pages={25688--25702},
  year={2021}
}

@article{hanzely2018sega,
  title={SEGA: Variance reduction via gradient sketching},
  author={Hanzely, Filip and Mishchenko, Konstantin and Richt{\'a}rik, Peter and A},
  journal={Advances in Neural Information Processing Systems},
  volume={31},
  year={2018}
}

@article{khirirat2018distributed,
  title={Distributed learning with compressed gradients},
  author={Khirirat, Sarit and Feyzmahdavian, Hamid Reza and Johansson, Mikael and B},
  journal={arXiv preprint arXiv:1806.06573},
  year={2018}
}

@inproceedings{gorbunov2021marina,
  title={MARINA: Faster non-convex distributed learning with compression},
  author={Gorbunov, Eduard and Burlachenko, Konstantin P and Li, Zhize and Richt{\'a}rik, Peter},
  booktitle={International Conference on Machine Learning},
  pages={3788--3798},
  year={2021},
  organization={PMLR}
}

@article{tyurin2022dasha,
  title={DASHA: Distributed nonconvex optimization with communication compression, optimal oracle complexity, and no client synchronization},
  author={Tyurin, Alexander and Richt{\'a}rik, Peter},
  journal={arXiv preprint arXiv:2202.01268},
  year={2022}
}

@article{li2020acceleration,
  title={Acceleration for compressed gradient descent in distributed and federated optimization},
  author={Li, Zhize and Kovalev, Dmitry and Qian, Xun and Richt{\'a}rik, Peter},
  journal={arXiv preprint arXiv:2002.11364},
  year={2020}
}

@article{li2021canita,
  title={CANITA: Faster rates for distributed convex optimization with communication compression},
  author={Li, Zhize and Richt{\'a}rik, Peter},
  journal={Advances in Neural Information Processing Systems},
  volume={34},
  pages={13770--13781},
  year={2021}
}

@article{cotter2011better,
  title={Better mini-batch algorithms via accelerated gradient methods},
  author={Cotter, Andrew and Shamir, Ohad and Srebro, Nati and Sridharan, Karthik},
  journal={Advances in neural information processing systems},
  volume={24},
  year={2011}
}

@inproceedings{allen2016even,
  title={Even faster accelerated coordinate descent using non-uniform sampling},
  author={Allen-Zhu, Zeyuan and Qu, Zheng and Richt{\'a}rik, Peter and Yuan, Yang},
  booktitle={International Conference on Machine Learning},
  pages={1110--1119},
  year={2016},
  organization={PMLR}
}

@inproceedings{scaman2017optimal,
  title={Optimal algorithms for smooth and strongly convex distributed optimization in networks},
  author={Scaman, Kevin and Bach, Francis and Bubeck, S{\'e}bastien and Lee, Yin Tat and Massouli{\'e}, Laurent},
  booktitle={international conference on machine learning},
  pages={3027--3036},
  year={2017},
  organization={PMLR}
}

@article{shi2015extra,
  title={Extra: An exact first-order algorithm for decentralized consensus optimization},
  author={Shi, Wei and Ling, Qing and Wu, Gang and Yin, Wotao},
  journal={SIAM Journal on Optimization},
  volume={25},
  number={2},
  pages={944--966},
  year={2015},
  publisher={SIAM}
}

@article{lee2015distributed,
  title={Distributed stochastic variance reduced gradient methods and a lower bound for communication complexity},
  author={Lee, Jason D and Lin, Qihang and Ma, Tengyu and Yang, Tianbao},
  journal={arXiv preprint arXiv:1507.07595},
  year={2015}
}

@article{mishchenko2019distributed,
  title={Distributed learning with compressed gradient differences},
  author={Mishchenko, Konstantin and Gorbunov, Eduard and Tak{\'a}{\v{c}}, Martin and Richt{\'a}rik, Peter},
  journal={arXiv preprint arXiv:1901.09269},
  year={2019}
}

@article{wang2018atomo,
  title={Atomo: Communication-efficient learning via atomic sparsification},
  author={Wang, Hongyi and Sievert, Scott and Liu, Shengchao and Charles, Zachary and Papailiopoulos, Dimitris and Wright, Stephen},
  journal={Advances in Neural Information Processing Systems},
  volume={31},
  year={2018}
}

@inproceedings{mishchenko202099,
  title={99\% of worker-master communication in distributed optimization is not needed},
  author={Mishchenko, Konstantin and Hanzely, Filip and Richt{\'a}rik, Peter},
  booktitle={Conference on Uncertainty in Artificial Intelligence},
  pages={979--988},
  year={2020},
  organization={PMLR}
}

@article{vogels2019powersgd,
  title={PowerSGD: Practical low-rank gradient compression for distributed optimization},
  author={Vogels, Thijs and Karimireddy, Sai Praneeth and Jaggi, Martin},
  journal={Advances in Neural Information Processing Systems},
  volume={32},
  year={2019}
}

@article{safaryan2019stochastic,
  title={On stochastic sign descent methods},
  author={Safaryan, Mher and Richt{\'a}rik, Peter},
  year={2019},
  publisher={arXiv}
}

@inproceedings{bernstein2018signsgd,
  title={signSGD: Compressed optimisation for non-convex problems},
  author={Bernstein, Jeremy and Wang, Yu-Xiang and Azizzadenesheli, Kamyar and Anandkumar, Animashree},
  booktitle={International Conference on Machine Learning},
  pages={560--569},
  year={2018},
  organization={PMLR}
}

@article{beznosikov2020biased,
  title={On biased compression for distributed learning},
  author={Beznosikov, Aleksandr and Horv{\'a}th, Samuel and Richt{\'a}rik, Peter and Safaryan, Mher},
  journal={arXiv preprint arXiv:2002.12410},
  year={2020}
}

@article{horvath2023stochastic,
  title={Stochastic distributed learning with gradient quantization and double-variance reduction},
  author={Horv{\'a}th, Samuel and Kovalev, Dmitry and Mishchenko, Konstantin and Richt{\'a}rik, Peter and Stich, Sebastian},
  journal={Optimization Methods and Software},
  volume={38},
  number={1},
  pages={91--106},
  year={2023},
  publisher={Taylor \& Francis}
}

@inproceedings{richtarik20223pc,
  title={3PC: Three point compressors for communication-efficient distributed training and a better theory for lazy aggregation},
  author={Richt{\'a}rik, Peter and Sokolov, Igor and Gasanov, Elnur and Fatkhullin, Ilyas and Li, Zhize and Gorbunov, Eduard},
  booktitle={International Conference on Machine Learning},
  pages={18596--18648},
  year={2022},
  organization={PMLR}
}

@article{wangni2018gradient,
  title={Gradient sparsification for communication-efficient distributed optimization},
  author={Wangni, Jianqiao and Wang, Jialei and Liu, Ji and Zhang, Tong},
  journal={Advances in Neural Information Processing Systems},
  volume={31},
  year={2018}
}

@inproceedings{shi2019distributed,
  title={A distributed synchronous SGD algorithm with global top-k sparsification for low bandwidth networks},
  author={Shi, Shaohuai and Wang, Qiang and Zhao, Kaiyong and Tang, Zhenheng and Wang, Yuxin and Huang, Xiang and Chu, Xiaowen},
  booktitle={2019 IEEE 39th International Conference on Distributed Computing Systems (ICDCS)},
  pages={2238--2247},
  year={2019},
  organization={IEEE}
}

@article{jiang2018linear,
  title={A linear speedup analysis of distributed deep learning with sparse and quantized communication},
  author={Jiang, Peng and Agrawal, Gagan},
  journal={Advances in Neural Information Processing Systems},
  volume={31},
  year={2018}
}

@inproceedings{horvoth2022natural,
  title={Natural compression for distributed deep learning},
  author={Horv{\'o}th, Samuel and Ho, Chen-Yu and Horvath, Ludovit and Sahu, Atal Narayan and Canini, Marco and Richt{\'a}rik, Peter},
  booktitle={Mathematical and Scientific Machine Learning},
  pages={129--141},
  year={2022},
  organization={PMLR}
}

@article{faghri2020adaptive,
  title={Adaptive gradient quantization for data-parallel sgd},
  author={Faghri, Fartash and Tabrizian, Iman and Markov, Ilia and Alistarh, Dan and Roy, Daniel M and Ramezani-Kebrya, Ali},
  journal={Advances in neural information processing systems},
  volume={33},
  pages={3174--3185},
  year={2020}
}

@inproceedings{wu2018error,
  title={Error compensated quantized SGD and its applications to large-scale distributed optimization},
  author={Wu, Jiaxiang and Huang, Weidong and Huang, Junzhou and Zhang, Tong},
  booktitle={International Conference on Machine Learning},
  pages={5325--5333},
  year={2018},
  organization={PMLR}
}

@article{wen2017terngrad,
  title={Terngrad: Ternary gradients to reduce communication in distributed deep learning},
  author={Wen, Wei and Xu, Cong and Yan, Feng and Wu, Chunpeng and Wang, Yandan and Chen, Yiran and Li, Hai},
  journal={Advances in neural information processing systems},
  volume={30},
  year={2017}
}

@inproceedings{andrew1979some,
  title={Some complexity questions related to distributed computing},
  author={Andrew, C-C Yao},
  booktitle={Proc. 11th STOC},
  pages={209--213},
  year={1979}
}

@InProceedings{pmlr-v70-scaman17a,
  title = 	 {Optimal Algorithms for Smooth and Strongly Convex Distributed Optimization in Networks},
  author =       {Kevin Scaman and Francis Bach and S{\'e}bastien Bubeck and Yin Tat Lee and Laurent Massouli{\'e}},
  booktitle = 	 {Proceedings of the 34th International Conference on Machine Learning},
  pages = 	 {3027--3036},
  year = 	 {2017},
  editor = 	 {Precup, Doina and Teh, Yee Whye},
  volume = 	 {70},
  series = 	 {Proceedings of Machine Learning Research},
  month = 	 {06--11 Aug},
  publisher =    {PMLR},
  url = 	 {https://proceedings.mlr.press/v70/scaman17a.html}
}

@article{newman1991private,
  title={Private vs. common random bits in communication complexity},
  author={Newman, Ilan},
  journal={Information processing letters},
  volume={39},
  number={2},
  pages={67--71},
  year={1991},
  publisher={Citeseer}
}

@inproceedings{seide20141,
  title={1-bit stochastic gradient descent and its application to data-parallel distributed training of speech dnns},
  author={Seide, Frank and Fu, Hao and Droppo, Jasha and Li, Gang and Yu, Dong},
  booktitle={Fifteenth annual conference of the international speech communication association},
  year={2014}
}

@inproceedings{tang20211,
  title={1-bit adam: Communication efficient large-scale training with adam’s convergence speed},
  author={Tang, Hanlin and Gan, Shaoduo and Awan, Ammar Ahmad and Rajbhandari, Samyam and Li, Conglong and Lian, Xiangru and Liu, Ji and Zhang, Ce and He, Yuxiong},
  booktitle={International Conference on Machine Learning},
  pages={10118--10129},
  year={2021},
  organization={PMLR}
}

@article{alistarh2017qsgd,
  title={QSGD: Communication-efficient SGD via gradient quantization and encoding},
  author={Alistarh, Dan and Grubic, Demjan and Li, Jerry and Tomioka, Ryota and Vojnovic, Milan},
  journal={Advances in neural information processing systems},
  volume={30},
  year={2017}
}

@inproceedings{Aji_2017,
	doi = {10.18653/v1/d17-1045},
  
	url = {https://doi.org/10.18653%2Fv1%2Fd17-1045},
  
	year = 2017,
	publisher = {Association for Computational Linguistics},
  
	author = {Alham Fikri Aji and Kenneth Heafield},
  
	title = {Sparse Communication for Distributed Gradient Descent},
  
	booktitle = {Proceedings of the 2017 Conference on Empirical Methods in Natural
		          Language Processing}
}

@article{lin2017deep,
  title={Deep gradient compression: Reducing the communication bandwidth for distributed training},
  author={Lin, Yujun and Han, Song and Mao, Huizi and Wang, Yu and Dally, William J},
  journal={arXiv preprint arXiv:1712.01887},
  year={2017}
}

@inproceedings{he2016deep,
  title={Deep residual learning for image recognition},
  author={He, Kaiming and Zhang, Xiangyu and Ren, Shaoqing and Sun, Jian},
  booktitle={Proceedings of the IEEE conference on computer vision and pattern recognition},
  pages={770--778},
  year={2016}
}

@misc{jin_accelerated_2017,
	title = {Accelerated {Gradient} {Descent} {Escapes} {Saddle} {Points} {Faster} than {Gradient} {Descent}},
	url = {http://arxiv.org/abs/1711.10456},
	urldate = {2022-10-31},
	publisher = {arXiv},
	author = {Jin, Chi and Netrapalli, Praneeth and Jordan, Michael I.},
	month = nov,
	year = {2017},
	note = {arXiv:1711.10456 [cs, math, stat]},
}

@article{sagun2016eigenvalues,
  title={Eigenvalues of the hessian in deep learning: Singularity and beyond},
  author={Sagun, Levent and Bottou, Leon and LeCun, Yann and A},
  journal={arXiv preprint arXiv:1611.07476},
  year={2016}
}

@article{freund2022convergence,
  title={When is the Convergence Time of Langevin Algorithms Dimension Independent? A Composite Optimization Viewpoint},
  author={Freund, Yoav and Ma, Yi-An and Zhang, Tong and A},
  journal={Journal of Machine Learning Research},
  volume={23},
  number={214},
  pages={1--32},
  year={2022}
}

@techreport{murty1985some,
  title={Some {NP}-complete problems in quadratic and nonlinear programming},
  author={Murty, Katta G and Kabadi, Santosh N},
  year={1985}
}

@book{nesterov2003introductory,
  title={Introductory lectures on convex optimization: A basic course},
  author={Nesterov, Yurii},
  volume={87},
  year={2003},
  publisher={Springer Science \& Business Media}
}

@article{nesterov_cubic_2006,
	title = {Cubic regularization of {Newton} method and its global performance},
	volume = {108},
	issn = {0025-5610, 1436-4646},
	url = {http://link.springer.com/10.1007/s10107-006-0706-8},
	doi = {10.1007/s10107-006-0706-8},
	language = {en},
	number = {1},
	urldate = {2022-09-15},
	journal = {Mathematical Programming},
	author = {Nesterov, Yurii and Polyak, B.T.},
	month = aug,
	year = {2006},
	pages = {177--205},
}

@inproceedings{dwork2006differential,
  title={Differential privacy},
  author={Dwork, Cynthia},
  booktitle={Automata, Languages and Programming: 33rd International Colloquium, ICALP 2006, Venice, Italy, July 10-14, 2006, Proceedings, Part II 33},
  pages={1--12},
  year={2006},
  organization={Springer}
}

@inproceedings{hsu2012random,
  title={Random design analysis of ridge regression},
  author={Hsu, Daniel and Kakade, Sham M and Zhang, Tong and A},
  booktitle={Conference on learning theory},
  pages={9--1},
  year={2012},
  organization={JMLR Workshop and Conference Proceedings}
}

@article{sagun2017empirical,
  title={Empirical analysis of the hessian of over-parametrized neural networks},
  author={Sagun, Levent and Evci, Utku and Guney, V Ugur and Dauphin, Yann and Bottou, Leon},
  journal={arXiv preprint arXiv:1706.04454},
  year={2017}
}

@inproceedings{ghorbani2019investigation,
  title={An investigation into neural net optimization via hessian eigenvalue density},
  author={Ghorbani, Behrooz and Krishnan, Shankar and Xiao, Ying and A},
  booktitle={International Conference on Machine Learning},
  pages={2232--2241},
  year={2019},
  organization={PMLR}
}

@article{brock2018large,
  title={Large scale GAN training for high fidelity natural image synthesis},
  author={Brock, Andrew and Donahue, Jeff and Simonyan, Karen and A},
  journal={arXiv preprint arXiv:1809.11096},
  year={2018}
}

@article{chang2008libsvm,
  title={LIBSVM data: Classification, regression, and multi-label},
  author={Chang, Chih-Chung},
  journal={http://www. csie. ntu. edu. tw/\~{} cjlin/libsvmtools/datasets/},
  year={2008}
}

@article{polyak1964some,
  title={Some methods of speeding up the convergence of iteration methods},
  author={Polyak, Boris T},
  journal={Ussr computational mathematics and mathematical physics},
  volume={4},
  number={5},
  pages={1--17},
  year={1964},
  publisher={Elsevier}
}

@article{lin2015universal,
  title={A universal catalyst for first-order optimization},
  author={Lin, Hongzhou and Mairal, Julien and Harchaoui, Zaid},
  journal={Advances in neural information processing systems},
  volume={28},
  year={2015}
}

\newpage
\appendix

\section{Acceleration of CORE-GD}\label{sec: acc}
In the optimization community, the momentum technique is used to accelerate convergence of Gradient Descent. We also design an accelerated CORE-based algorithm. We name the algorithm as CORE-Accelerated Gradient Descent (CORE-AGD). Here we simply consider the objective function to be quadratic, i.e.
\begin{equation}\label{eq:quad}
  f(\x) = \frac{1}{2}\x^\top \A \x.  
\end{equation}
 This corresponds to picking $\sigma_i$ as quadratic in linear models. We also note that
the quadratic function is already very representative, as it is known that most worst-case  functions (lower-bound instances) in the convex optimization are exactly quadratic  (see e.g. \textcite[Chapter 2]{nesterov2003introductory}).  As we have mentioned, our analysis can be directly extended to general convex optimization under an additional Hessian smoothness condition combining with higher-order methods.  We present our algorithm in Algorithm \ref{alg:CORE-AGD}, which is a heavy-ball \cite{polyak1964some} based algorithm by replacing the gradient to be its estimation using CORE.  

\begin{algorithm*}[t]
    \caption{CORE-AGD}\label{alg:CORE-AGD}
    \begin{algorithmic}
    \REQUIRE{$n$ machines, a central machine, a common random number generator, $m\le \frac{\tr(\A)}{L}$, $x^0$, $k=0$, $\beta \gets \sqrt{h\mu},h\gets\frac{m^2}{14400^2(\sum_i\lambda_i^{1/2})^2}$}
    \WHILE{$k\leq N$}
        \STATE{Generate fresh i.i.d. $m$ Gaussian vectors $\gauss_1,\cdots,\gauss_m$ with the common random number generator.}
    \STATE{Machine $i$ computes $\y^k = \x^k+(1-\beta)(\x^k-\x^{k-1}) $ and sends $p_{ij}=\langle\nabla f_i(\y^k),\gauss_j \rangle$ to the central machine.} 
    \STATE{The central machine sends $\sum_{i=1}^n p_{ij}$ back to every machine.}
    \STATE{Machines reconstruct $\tilde \nabla_m f(\y^k)$ by
    \begin{equation}
        \tilde \nabla_m f(\x^k) = \frac{1}{m}\sum_{i=1}^n\sum_{j=1}^m p_{ij}\gauss_j
    \end{equation}}
    \STATE{Machines update $\x_k$  by $\tilde \nabla_m f(\y^k)$ as
    \begin{equation}
    \begin{split}
        \x^{k+1} = \y^k-h\tilde\nabla (\y^k)
        \end{split}
    \end{equation}}
    \ENDWHILE
    \end{algorithmic}
\end{algorithm*}
By a careful analysis of the CORE-AGD algorithm, we have the following theorem:
\begin{theorem}
For objective in form of \eqref{eq:quad}, let $\{\lambda_i\}_{i=1}^d$ be the eigenvalues of $\A$ with a decreasing order, and denote $L=\lambda_1$, $\mu=\lambda_d$. Under the hyper-parameter setting in Algorithm \ref{alg:CORE-AGD},  we have
    \begin{equation}
        \E f(\x^N) \le 400\left(1-\frac{1}{57600} \frac{m\mu^{1/2}}{\sum_{i=1}^d \lambda_i^{1/2}} \right)^{N}\cdot \frac{L}{\mu}\cdot f(\x^0).
    \end{equation}
    \label{thm:CORE-AGD}
\end{theorem}
In Theorem \ref{thm:CORE-AGD}, if $f$ is not strongly-convex ($\lambda_d=0$) or $\lambda_d$ is too small, i.e. ($\lambda<\epsilon$), we can also use the reduction technique (see e.g. \cite{lin2015universal}) by
adding a regularization term.
From Theorem \ref{thm:CORE-AGD},
the total communication costs to find an $\epsilon$-approximate solution for CORE-AGD are $\tilde{\cO}\left(\frac{\sum_{i=1}^d \lambda_i^{1/2}}{\mu^{1/2}}\right)$. In contrast, the communication costs of CAGD are $\tilde{\cO}\left(\frac{d\lambda_1^{1/2}}{\mu^{1/2}}\right)$. Again, we obtain a provably better communication costs because $\frac{\sum_{i=1}^d \lambda_i^{1/2}}{\mu^{1/2}} \le \frac{dL^{1/2}}{\mu^{1/2}}$ when ignoring logarithmic factors. And when $m = \Theta\left(\frac{\sum_{i=1}^d \lambda_i^{1/2}}{L^{1/2}}\right)$,  CORE-AGD achieves the same number of communication rounds (convergence rate) as those of CAGD when ignoring logarithmic factors. We then specify the objective to satisfy the ridge-separable form \eqref{eq: ridgeseparable} with $\sigma_i$ being a quadratic function. We have  Corollary \ref{cor:linear-model acc}, which states that CORE-AGD  reduces the communication costs by  a  $\sqrt{\min(d,\alpha^{-1})}$ factor compared with the ``worst-case-optimal'' CAGD algorithm.
\begin{corollary}
For the objective function in form of  \eqref{eq: ridgeseparable} with $\sigma_i$ being a quadratic function, under Assumptions \ref{asp: data} with $R$ treated as a constant,   
the total communication costs of CORE-AGD are $\tilde\cO\left(d+\frac{\sqrt{dL_0R}}{\alpha} \right)$. 
\label{cor:linear-model acc}
\end{corollary}

\section{Decentralized CORE Based Algorithms}
\label{sec:decentralized}
In this section, we consider the decentralized optimization settings. In centralized settings, we assume that all the machines can send the gradient to the central machine. However, if machines can only send messages to their neighbours, a message will be transmitted several times before it reaches the central machine. In the worst case, the total communication costs will be multiplied by the diameter of the graph. In the decentralized settings, the communication costs usually depend on the gossip matrix $\W$ of the graph. We propose decentralized CORE-GD in Algorithm \ref{alg:CORE-GD-decentralized}, and analyze its communication costs.

\begin{algorithm*}[t]
\caption{Decentralized CORE-GD with per-round communication budget $m$}\label{alg:CORE-GD-decentralized}
\begin{algorithmic}
\REQUIRE{$n$ machines, a central machine, a common random number generator, $m\le \frac{\tr(\A)}{L}$, $\x^0$, $k=0$,  step-size $h_k = \frac{m}{4\tr(\A)}$.}
\WHILE{$k<N$}
    \STATE{Generate fresh i.i.d. $m$ Gaussian vectors $\gauss_1,\cdots,\gauss_m$ with the common random number generator.}
    \STATE{Machine $i$ computes projections $p_{ij}=\langle\nabla f_i(\x^k),\gauss_j \rangle$ locally. Define $\p_i = \begin{bmatrix}
        p_{i1}&\cdots&p_{im}
    \end{bmatrix}^\top$.} 
    \STATE{Machines solve an $m$-dimensional subproblem with an decentralized optimization algorithm:
    \begin{equation}
        \p = \argmin_{\x\in \R^m} \frac{1}{n}\sum_{i=1}^{n} \frac{1}{2}\|\x-\p_{i}\|^2.
    \end{equation}}
    \STATE{Denote $p_j$ to be the $j$th coordinate of $\p$. Machines reconstruct $\tilde \nabla_m f(\x^k)$ by
    \begin{equation}
        \tilde \nabla_m f(\x^k) = \frac{n}{m}\sum_{j=1}^m p_{j}\gauss_j,
        \label{equ:subproblem}
    \end{equation}
    }
    \STATE{Machines update $\x^k$ by \begin{equation}
    \begin{aligned}
    \x^{k+1} &= \x^k - h_k\tilde\nabla_m f(\x^k).
    \end{aligned}
    \end{equation}}
    \STATE{$k\gets k+1$.}
\ENDWHILE
\end{algorithmic}
\end{algorithm*}

The optimal solution of subproblem \eqref{equ:subproblem} is
\begin{equation}
    \p = \frac{1}{n} \sum_{i=1}^n \p_i.
\end{equation}
Therefore, we have
\begin{equation}
p_j = \frac{1}{n}\sum_{i=1}^n p_{ij}.
\end{equation}
By solving supproblem \eqref{equ:subproblem}, we broadcast $p_j$ to every machine in the graph, and each machine can reconstruct the gradient using $p_j$. The Hessian matrix of the objective function in \eqref{equ:subproblem} is $\I_m$, so \eqref{equ:subproblem} is simple to optimize. GD will find the optimal solution in one step if \eqref{equ:subproblem} can be solved locally. The optimal communication costs of solving \eqref{equ:subproblem} to accuracy $\epsilon$ is $\cO\left(\frac{1}{\sqrt{\gamma}}\log\frac{1}{\epsilon}\right)$, where $\gamma$ is the eigengap of the gossip matrix $\W$ of the graph (see e.g. \textcite{pmlr-v70-scaman17a}). Ignoring logarithmic factors, the total communication costs of decentralized CORE-GD are only $\tilde\cO\left(\frac{1}{\sqrt{\gamma}}\right)$ times more than the communication costs of centralized CORE-GD in the same setting.

\section{Deferred Proofs in Section \ref{sec:core idea} }
\begin{proof}[Proof of Lemma \ref{lem:unbiased}]
\begin{equation}
    \begin{aligned}
        \E_{\gauss_1,\cdot,\gauss_m} \tilde\ba &= \E_{\gauss_1,\cdots,\gauss_m} \left[ \frac{1}{m}\sum_{i=1}^m \langle\ba,\gauss_i\rangle\cdot\gauss_i\right]\\
        &= \E_{\gauss_1} \gauss_{1} \gauss_{1}^{\top} \ba = \I \ba\\
        &=  \ba
    \end{aligned}
\end{equation}
\end{proof}

\begin{proof}[Proof of Lemma \ref{lem:descent}]
For the simplicity of notation, we use $\E_\gauss$ to denote  $\E_{\gauss_1,\cdots,\gauss_m}$.
\begin{equation}
    \begin{aligned}
        \E_\gauss \|\tilde\ba - \ba \|_\A^2 &= \E_{\gauss} \left\|\frac{1}{m}\sum_{i=1}^m(\langle\ba, \gauss\rangle\cdot \gauss - \ba)\right\|_\A^2\\
        &= \E_\gauss \left[\frac{1}{m^2} \sum_{i=1}^m \left(\ba^\top\gauss_i\gauss_i^\top\A\gauss_i\gauss_i^\top \ba - \ba^\top\A\ba\right)\right]\\
        &= \frac{1}{m} \E_{\gauss_1} \ba^\top\gauss_1\gauss_1^\top\A\gauss_1\gauss_1^\top\ba - \frac{1}{m} \|\ba\|_\A^2.
    \end{aligned}
    \label{equ:lemdescent1}
\end{equation}
Let $\A = \U^\top\D \U$ be the eigenvalue decomposition of $\A$ where $\D = \mathrm{diag}\{b_1,\cdots,b_d\}$ is a diagonal matrix, and $\boldsymbol{\zeta}=\U\gauss_1$ be a linear transformation of the random variable $\gauss_1$. We have
\begin{equation}
    \begin{aligned}
        \E_{\gauss_1} \left[\gauss_1\gauss_1^\top\A\gauss_1\gauss_1^\top \right] &\overset{a}= \E_{\boldsymbol{\zeta}} \left[\U^\top\gz\gz^\top\D\gz\gz^\top\U \right]\\
        &= \U^\top\E_\gz\left[\sum_{i=1}^d b_i\gz_i^2 \cdot\gz\gz^\top \right]\U\\
        &\overset{b}= \U^\top\left(\sum_{i=1}^d b_i\cdot \I + 2\D \right)\U\\
        &\overset{c}= \tr(\A)\cdot \I + 2\A\\
        &\preceq 3\tr(\A)\cdot \I.
    \end{aligned}
    \label{equ:lemdescent2}
\end{equation}
In $\overset{a}=$, we use $\gz \sim N(0,\I_d)$ based on the rotational invariance of the standard Gaussian distribution. In $\overset{b}=$, we use the second and forth moment of standard Gaussian variables: $\E\gz_i^2 = 1$ and $\E\gz_i^4 = 3$. In $\overset{c}=$, we use $\tr(\U^\top\D\U) = \tr(\U^\top\U\D) = \tr(\D)$. The last inequality of \eqref{equ:lemdescent2} is due to $\tr(\A)\cdot \I \succeq \A$. Combining \eqref{equ:lemdescent2} and \eqref{equ:lemdescent1}, we have
\begin{equation}
    \E_{\gauss_1,\cdots,\gauss_m} \|\tilde \ba - \ba \|_\A^2 \le \frac{3\tr(\A)}{m}\|\ba\|^2 - \frac{1}{m}\|\ba\|_\A^2.
    \label{equ:lemdescent3}
\end{equation}
\end{proof}

\section{Deferred Proofs in Section \ref{sec: linear model} }
\begin{proof}[Proof of Theorem \ref{thm:CORE-GD-sconvex}]
We write the second-order Taylor expansion of $f(\x^{k+1})$ at $\x^k$:
\begin{equation}
\begin{split}
    f(\x^{k+1}) &\le f(\x^k) + \langle\nabla f(\x^k), \x^{k+1}-\x^k\rangle + \frac12\langle \A (\x^{k+1}-\x^k), \x^{k+1}-\x^k\rangle.
\end{split}
    \label{equ:secondTaylor1}
\end{equation}
Combining the updating process of $\x^{k+1}$ and \eqref{equ:secondTaylor1}, we have
\begin{equation}
    \begin{split}
    f(\x^{k+1}) &\le f(\x^k) - h_k\langle\nabla f(\x^k), \tilde\nabla_m f(\x^k)\rangle + \frac{h_k^2}{2}\|\tilde\nabla_m f(\x^k)\|_{\A}^2.
    \end{split}
    \label{equ:secondTaylor2}
\end{equation}
Taking expectation with respect to $\tilde \nabla_m f(\x^k)$ to both sides of \eqref{equ:secondTaylor2}, using Lemma \ref{lem:unbiased}, Lemma \ref{lem:descent} and Definition \ref{AH}, we have 
\begin{equation}
\begin{split}
    \E f(\x^{k+1}) &\le f(\x^k) - h_k\|\nabla f(\x^k)\|^2 + h_k^2\left(\frac{3\tr(\nabla^2 f(\x^k))}{2m} \|\nabla f(\x^k)\|^2 + \|\nabla f(\x^k)\|_\A^2\right) \\
    &\le f(\x^k) - \left(h_k- h_k^2\left(\frac{3\tr(\A)}{2m} + L\right)\right) \|\nabla f(\x^k)\|^2. \\
    &\overset{a}\le f(\x^k) - \left(h_k- h_k^2\cdot\frac{5\tr(\A)}{2m} \right) \|\nabla f(\x^k)\|^2,
    \label{equ:secondex}
\end{split}
\end{equation}
where in $\overset{a}\le$ we use $m\le \frac{\tr(\A)}{L}$. Then, using $\mu$-strongly convex condition, we have
\begin{equation}
\begin{split}
    f^* &\geq \min_{\y} \left\{ f(\x^k)+\langle \nabla f(\x^k), \y-\x^k \rangle +\frac{\mu}{2}\Vert \y-\x^k\Vert^2 \right\}\\
    & = f(\x^k) - \frac{1}{2\mu}\Vert \nabla f(\x^k)\Vert^2.
    \label{equ:con}
\end{split}
\end{equation}
Combining \eqref{equ:secondex} and \eqref{equ:con}, we have
\begin{equation}
\begin{split}
    \E f(x^{k+1})-f^* &\le f(\x^k)-f^* - 2\mu\left(h_k- \frac{5h_k^2\tr(\A)}{2m}\right)\left(f(\x^k)-f^*\right)\\
    &\overset{a}= \left(1-\frac{3m\mu}{16\tr(\A)}\right)\left(f(\x^k)-f^*\right),\label{le}
\end{split}
\end{equation}
where in $\overset{a}=$ we use $h_k = \frac{m}{4\tr(\A)}$. Thus, we finish the proof of Theorem \ref{thm:CORE-GD-sconvex}.
\end{proof}

\section{Deferred Proofs in Section \ref{sec:non-convex}}
\setcounter{theorem}{1} 
\renewcommand{\thetheorem}{5.\arabic{theorem}}
In this section, we prove Theorem \ref{thm:CORE-GDAS} as below. 
\begin{theorem}
    Assume that $f(\x)$ is $L$-smooth and has $H$-Lipschitz continuous Hessian matrix. With the assumption of $\tr(\nabla^2 f(\x))\le \effdim_1$ for any $\x\in\R^d$ and $f(\x^0)-f^*\le\Delta$. Then, under the hyper-parameter setting in Algorithm \ref{alg:CORE-GDAS}, the following result in expectation 
    \begin{equation}
        \E f(\x^k) \le f(\x^0) - \sum_{i=1}^k \E \left[ \frac{h_i}{2}\|\nabla f(\x^i)\|^2\right]
    \end{equation}
    holds for option II, and holds with probability $1-\delta$ for option I.
    \label{thm:CORE-GDAS1}
\end{theorem}

\begin{proof}[Proof of Theorem \ref{thm:CORE-GDAS1}]
We write the third-order Taylor expansion of $f(\tilde \x^{k+1})$ at $\x^k$:
\begin{equation}
\begin{split}
    f(\tilde\x^{k+1}) &\le f(\x^k) + \langle\nabla f(\x^k), \tilde\x^{k+1}-\x^k\rangle + \frac12\langle \nabla^2 f(\x^k) (\tilde\x^{k+1}-\x^k), \tilde\x^{k+1}-\x^k\rangle \\
    &+ \frac{H}{6} \|\tilde\x^{k+1}-\x^k\|^3.
    \end{split}
    \label{equ:thirdTaylor1}
\end{equation}
Combining the updating process of $\tilde\x^{k+1}$ with \eqref{equ:thirdTaylor1}, we have
\begin{equation}
\begin{split}
     f(\tilde\x^{k+1}) &\le f(\x^k) - h_k\langle\nabla f(\x^k), \tilde\nabla_m f(\x^k)\rangle+ \frac{h_k^2}{2}\|\tilde\nabla_m f(\x_k)\|_{\nabla^2 f(\x^k)}^2 + \frac{Hh_k^3}{6} \|\tilde\nabla_m f(\x_k)\|^3.
     \end{split}
    \label{equ:thirdTaylor2}
\end{equation}
We denote $\E_k[\cdot] = \E[\cdot |x_k]$. Then taking expectation with respect to $\tilde \nabla_m f(\x^k)$ to both sides of \eqref{equ:thirdTaylor2} and using Lemma \ref{lem:unbiased} and Lemma \ref{lem:descent}, we have
\begin{equation}
\begin{split}
    \E_k f(\tilde \x^{k+1}) &\le f(\x^k) - h_k\|\nabla f(\x^k)\|^2 + \frac{3h_k^2\tr(\nabla^2 f(\x^k))}{2m} \|\nabla f(\x^k)\|^2 + \frac{Hh_k^3}{6}\E_k \|\tilde\nabla_m f(\x_k)\|^3\\
    &\le f(\x^k) - h_k\|\nabla f(\x^k)\|^2 + \frac{3h_k^2\effdim_1}{2m} \|\nabla f(\x^k)\|^2 + \frac{Hh_k^3}{6}\E_k \|\tilde\nabla_m f(\x^k)\|^3.
\end{split}
\label{equ:thirdTaylor3}
\end{equation}

Now we give an upper bound of $\E_k \|\tilde \nabla_m f(\x^k)\|^3$. We suppose the $m$ random Gaussian vectors are $\gauss_i$ for $i\in \{1,\cdots,m\}$. And we denote each $\gauss_i$ as 
\begin{equation}
\gauss_i = \begin{bmatrix}\xi_{i1}\\\vdots\\\xi_{id}\end{bmatrix},
\end{equation}
where $\xi_{ij}\sim N(0,1)$ is independent to each other. Then we have
\begin{equation}
\begin{split}
    \E_k \|\tilde \nabla_m f(\x^k)\|^3
    &\le \left( \E_k \|\tilde \nabla_m f(\x^k)\|^6 \right)^{1/2}\\
    &\overset{a}\le \left( 64\|\nabla f(\x^k)\|^6 + 64\E_k \left\|\tilde \nabla_m f(\x^k) - \nabla f(\x^k)\right\|^6  \right)^{1/2}\\
    &\overset{b}= 8\|\nabla f(\x^k)\|^3 \cdot\left(1+\E\left(\left(\frac{1}{m}\sum_{i=1}^m(\gz_{i1}^2-1) \right)^2 + \sum_{j=2}^d \left(\frac{1}{m}\sum_{i=1}^m\gz_{i1}\gz_{ij} \right)^2  \right)^3 \right)^{1/2}\\
    &\overset{c}\le 8\|\nabla f(\x^k)\|^3\left(1+20000\frac{d^3}{m^3} \right)^{1/2}\\
    &\le 1600\frac{d^{3/2}}{m^{3/2}}\|\nabla f(\x^k)\|^3.
\end{split}
\label{equ:thirdexp}
\end{equation}
In $\overset{a}\le$, we use the upper bound of the sixth moment as below.
\begin{equation}
\begin{split}
    \E\left\|X\right\|^6 &\leq \E\left(\left\| X-\E X\right\| + \left\|\E X\right\|\right)^6 \\
    &\leq\E \left(2\max\left\{ \left\| X-\E X\right\|, \left\|\E X\right\|\right\}\right)^6\\
    &\leq 64\E\left\|X -\E X\right\|^6 + 64 \left\|\E X\right\|^6. 
\end{split}
\end{equation}
In $\overset{b}=$, we analyse $\left\|\tilde \nabla_m f(\x^k) - \nabla f(\x^k)\right\|^2$ as below. Considering the rotation invariance of the standard Gaussian vectors, we can simplify the computation by rotating the coordinate system. For simplicity, we denote $\nabla f(\x^k) = \ba$. We can find an orthogonal matrix $\U$ such that $\U\ba = \begin{bmatrix}
    \|\nabla f(\x^k)\|,0,\cdots,0
\end{bmatrix}^\top$. Letting $\hat\ba = \U\ba $ and $\mathbf{\gz}_i = \U\gauss_i$, we have $\mathbf{\gz}_i \sim N(0, \I_d)$ and we denote $\mathbf{\gz}_i$ as
\begin{equation}
    \mathbf{\gz}_i = \begin{bmatrix}\gz_{i1}\\\vdots\\\gz_{id}\end{bmatrix},
\end{equation}
where $\gz_ij \sim N(0,1)$ is also independent to each other. Then we have
\begin{equation}
\begin{split}
    \left\|\tilde \nabla_m f(\x^k) - \nabla f(\x^k)\right\|^2 &= \left \| \U\left(\tilde \nabla_m f(\x^k) - \nabla f(\x^k)\right) \right\|^2
    = \left\|\frac{1}{m}\sum_{i=1}^m \left(\ba^\top\gauss_i \U\gauss_i -\U\ba \right) \right\|^2 \\
    &= \left\|\frac{1}{m}\sum_{i=1}^m \left((\U\ba)^\top(\U\gauss_i) \U\gauss_i -\U\ba \right) \right\|^2 \\
    &= \left\|\frac{1}{m}\sum_{i=1}^m \left(\hat\ba^\top\mathcal{\gz}_i \mathcal{\gz}_i -\hat\ba \right) \right\|^2\\
    &=\left\|\nabla f(\x^k)\right\|^2\left(\left(\frac{1}{m}\sum_{i=1}^m(\gz_{i1}^2-1) \right)^2 + \sum_{j=2}^d \left(\frac{1}{m}\sum_{i=1}^m\gz_{i1}\gz_{ij} \right)^2  \right). \nonumber
\end{split}
\end{equation}

In $\overset{c}\le$, we calculate the high-order moment of standard Gaussian distribution. Especially, we have
$\E\gz_i^{2n} = O(1)$ and $\E\gz_i^{2n+1} = 0$, where $n \in \{1,2,3,4,5,6\}$ ensuring that
\begin{equation}
    \E\left(\left(\frac{1}{m}\sum_{i=1}^m(\gz_{i1}^2-1) \right)^2 + \sum_{j=2}^d \left(\frac{1}{m}\sum_{i=1}^m\gz_{i1}\gz_{ij} \right)^2  \right)^3 = O\left(\frac{d^3}{m^3}\right).
\end{equation}

Combining \eqref{equ:thirdexp} with \eqref{equ:thirdTaylor3}, we have 
\begin{equation}
\begin{split}
    \E_k f(\tilde \x^{k+1}) &\le f(\x^k) - h_k\|\nabla f(\x^k)\|^2 + \frac{3h_k^2\effdim_1}{2m} \|\nabla f(\x^k)\|^2 + \frac{800Hh_k^3d^{3/2}}{3m^{3/2}}\|\nabla f(\x^k)\|^3.
    \end{split}
    \label{summing}
\end{equation}

For option I, we define the event
\begin{equation}
    \mathcal{H}_{N} = \left( p \geq 2\|\nabla f(\x^k)\|, \quad \forall k \leq N-1 \right).
\end{equation}
Let the event $\tilde{\mathcal{H}}_k = \left( p \geq 2\|\nabla f(\x^k)\|\right)$, with $0 \leq k \leq N-1$. Our choice of $m$ ensures for all $0 \leq k \leq N-1$, $\tilde{\mathcal{H}}_k$ occurs with probability at least $1 - \frac{\delta}{N}$. So, we have
\begin{equation}
    \mathbf{P}\left( \mathcal{H}_{N} \right) = \mathbf{P}\left( \bigcap_{k=0}^{N-1} \tilde{\mathcal{H}}_k \right) \geq 1- \sum_{k=0}^{N-1} \mathbf{P}\left(\tilde{\mathcal{H}}_k^c \right) \geq 1-\delta.
\end{equation}

Therefore, with probability at least $1-\delta$, $p \geq 2\|\nabla f(\x^k)\|$ holds for all $k=1,\cdots,N-1$. In the high-probability case, we have
\begin{equation}
    \frac{3h_k^2\effdim_1}{2m} \|\nabla f(\x^k)\|^2\le \frac{1}{4}h_k\|\nabla f(\x^k)\|^2,
    \label{equ:GDAS11}
\end{equation}
and 
\begin{equation}
    \frac{800Hh_k^3d^{3/2}}{3m^{3/2}}\|\nabla f(\x^k)\|^3\le \frac{1}{4}h_k\|\nabla f(\x^k)\|^2.
    \label{equ:GDAS22}
\end{equation}

For option II, By the choice of $h_k$, we also have 
\begin{equation}
    \frac{3h_k^2\effdim_1}{2m} \|\nabla f(\x^k)\|^2\le \frac{1}{4}h_k\|\nabla f(\x^k)\|^2,
    \label{equ:GDAS1}
\end{equation}
and 
\begin{equation}
    \frac{800Hh_k^3d^{3/2}}{3m^{3/2}}\|\nabla f(\x^k)\|^3\le \frac{1}{4}h_k\|\nabla f(\x^k)\|^2.
    \label{equ:GDAS2}
\end{equation}

Therefore, by summing the \eqref{summing} over $k$ and taking the full expectation, we have
    \begin{equation}
        \E f(\x^k)  \le f(\x^0) - \sum_{i=1}^k \E \left[ \frac{h_i}{2}\|\nabla f(\x^i)\|^2\right],
    \end{equation}
which holds with probability $1-\delta$ for option I and holds for option II. Now we take a deeper discussion.

\begin{itemize}
    \item For Option I,  In the high-probability case, in $N$ iterations, there are at least $N/2$ rounds of $h_k = \frac{m}{16\effdim_1(f)}$ or $N/2$ rounds of $H^{-1/2}p^{-1/2}d^{-3/4}m^{3/4}$, and in every round $\E f(\x^k)$ decreases by $\E\left[\frac{h_k}{2} \|\nabla f(\x^k)\|^2\right]$. Therefore, CORE-GD needs $\cO\left(\max\left\{\frac{\Delta\effdim_1(f)}{m\epsilon^2}, \frac{\Delta H^{1/2}d^{3/4}}{m^{3/4}\epsilon^{3/2}} \right\}\right)$ rounds to find an $\epsilon$-stationary point from $\{\x^k\}_{k=0}^{N-1} $ with probability $1-\delta$. The total communication costs of CORE-GD are $\cO\left(\max\left\{\frac{\Delta\effdim_1(f)}{\epsilon^2}, \frac{\Delta H^{1/2}d^{3/4}m^{1/4}}{\epsilon^{3/2}} \right\}\right)$.
    \item For Option II, in $N$ iterations, there are at least $N/2$ rounds of $h_k = \frac{m}{16\effdim_1(f)}$ or $N/2$ rounds of $H^{-1/2}(L\Delta)^{-1/4}d^{-3/4}m^{3/4}$, and in every round $\E f(\x^k)$ decreases by $\E\left[\frac{h_k}{2} \|\nabla f(\x^k)\|^2\right]$. Therefore, CORE-GD needs $\cO\left(\max\left\{\frac{\Delta\effdim_1(f)}{m\epsilon^2}, \frac{\Delta^{5/4}L^{1/4} H^{1/2}d^{3/4}}{m^{3/4}\epsilon^{2}} \right\}\right)$ rounds to find an $\epsilon$-stationary point from $\{\x^k\}_{k=0}^{N-1} $ in high probability. The total communication costs of CORE-GD are $\cO\left(\max\left\{\frac{\Delta\effdim_1(f)}{\epsilon^2}, \frac{\Delta^{5/4}L^{1/4} H^{1/2}d^{3/4}m^{1/4}}{\epsilon^{2}} \right\}\right)$.
\end{itemize}

\end{proof}

\section{Deferred Proofs in Appendix \ref{sec: acc}}
\begin{proof}[Proof of Theorem \ref{thm:CORE-AGD}]
Before our proof, we propose a useful Lemma taken from \cite{jin_accelerated_2017}.
\begin{lemma}
    Let the $2\times 2$ matrix $\A$ have following form, for arbitrary $a,b \in \R$,
    \begin{equation}
        \A = \begin{bmatrix}
            a&b\\
            1&0
        \end{bmatrix}.
    \end{equation}
    Letting $\mu_1$, $\mu_2$ denote the two eigenvalues of $\A$, then, for any $t \in \N$,
    \begin{equation}
        \begin{split}
            &\begin{bmatrix}
                1&0
            \end{bmatrix}\A^t = \left(\sum_{i=0}^t \mu_1^i\mu_2^{t-i} \qquad -\mu_1\mu_2\sum_{i=0}^{t-1} \mu_1^i\mu_2^{t-i-1} \right),\\
            &\begin{bmatrix}
                0&1
            \end{bmatrix}\A^t = \begin{bmatrix}
                1&0
            \end{bmatrix}\A^{t-1}.
        \end{split}
    \end{equation}
    \label{lemmajin}
\end{lemma}
Now we start our proof. Let $\z^{k+1} = \begin{bmatrix}
        \x^{k+1}\\
        \x^k
    \end{bmatrix}$. The iterations of CORE-AGD can be written as
    \begin{equation}
        \z^{k+1} = \begin{bmatrix}
            (2-\beta)(\I-h\A)&-(1-\beta)(\I-h\A)\\
            \I&\mathbf 0
        \end{bmatrix}\z^k+h\beps^k
        \stackrel{\triangle}{=}\B\z^k+h\beps^k,\\
        \label{equ:zdef}
    \end{equation}
    
    where $\beps^k = \begin{bmatrix}(\I-\frac{1}{m}\sum_{i=1}^m \gauss_i\gauss_i^\top)\A\y_k\\\mathbf 0\end{bmatrix}$, representing the error of estimating $\nabla f(\x_k)$ with $\tilde\nabla_m f (\x_k)$. 

    By induction on $k$, we have
    \begin{equation}
        \z^{N} = \B^N \z^0 + h\sum_{k=0}^{N-1} \B^{N-k-1}\beps^{k}.
        \label{equ:recursive}
    \end{equation}

    Without loss of generality, we assume that $\x^*=\mathbf 0$. We estimate the distance to the optimal solution by the $\A^2$ norm of $\x^k$. To compute $\|\x^k\|_{\A^2}$, we decompose $\x^k$ into eigen-directions of $\A$, and $\B$ can be decomposed into $2\times2$ matrices. For an eigen-direction with eigenvalue $\lambda$, the update of $\AGD$ can be written as follows:
    
    \begin{equation}
    \begin{aligned}
        \begin{bmatrix}
            x_{k+1}\\x_k
        \end{bmatrix} &= 
        \begin{bmatrix}
            (2-\beta)(1-h\lambda)&-(1-\beta)(1-h\lambda)\\
            1&0
        \end{bmatrix}\begin{bmatrix}
            x_{k}\\x_{k-1}
        \end{bmatrix}
        +h\begin{bmatrix}
            \epsilon\\0
        \end{bmatrix}\\
        &\stackrel{\triangle}{=}\B_\lambda\begin{bmatrix}
            x_{k}\\x_{k-1}
        \end{bmatrix}
        +h\begin{bmatrix}
            \epsilon\\0
        \end{bmatrix}.
    \end{aligned}
    \end{equation}
    Let $\mu_1$ and $\mu_2$ be the eigenvalues of $\B_\lambda$. 
Let $\C = \begin{bmatrix}
        \A^2&\mathbf 0\\\mathbf 0&\mathbf \A^2
    \end{bmatrix}$. By \eqref{equ:recursive}, We have

    \begin{equation}
        \begin{aligned}
            \E \|\z^{N}\|_\C^2
            &\le 2\|\B^N\z^0\|_{\C}^2 +2\E \left\|\sum_{k=0}^{N-1} \B^{N-k-1} \beps^k\right\|_\C^2.\\
        \end{aligned}
        \label{equ:znexpansion}
    \end{equation}
    
    For the $\beps^k$ terms, we have 
    \begin{equation}
        \begin{aligned}
            &\quad\E \left\|\sum_{k=0}^{N-1} \B^{N-k-1} \beps^k \right\|_\C^2 \\
            &= \sum_{k=0}^{N-1} \E_\gauss \left\| \B^{N-k-1} \beps^k \right\|_\C^2\\
            &= \sum_{k=0}^{N -1}\E_\gauss \begin{bmatrix}
                {\y^k}^\top\A^\top(\I-\frac{1}{m}\sum_{j=1}^m \gauss_j\gauss_j^\top)&\mathbf 0
            \end{bmatrix} (\B^{N-k-1})^\top \C \B^{N-k-1}\begin{bmatrix}
                (\I-\frac{1}{m}\sum_{j=1}^m \gauss_j\gauss_j^\top)\A\y^k\\\mathbf 0
            \end{bmatrix}\\
            &\stackrel{\text{Lemma \ref{lem:descent}}}{\le} 3\sum_{k=0}^{N-1} \tr\left((\B^{N-k-1})^\top \C \B^{N-k-1}\right)\cdot \frac{\|\y^k\|_{\A^2}^2}{m}.
        \end{aligned}
    \end{equation}
    In order to estimate $\tr\left((\B^{N-k-1})^\top \C \B^{N-k-1}\right)$, we consider blocks of $\B$ with respect to eigen-directions of $\A$. The contribution of an eigen-direction with eigenvalue $\lambda$ in the trace is
    \begin{equation}
        \begin{aligned}
        &\quad\tr \left((\B_\lambda^{N-k-1})^\top\cdot\begin{bmatrix} \lambda^2&0\\0&\lambda^2\end{bmatrix} \B_\lambda^{N-k-1}\right)\\
        &= \lambda^2\left(\left\|\begin{bmatrix}
            1&0
        \end{bmatrix}\B_\lambda^{N-k-1}\right\|^2 + \left\|\begin{bmatrix}
            0&1
        \end{bmatrix}\B_\lambda^{N-k-1}\right\|^2\right)
        \end{aligned}
        \label{equ:trace}
    \end{equation}
    By Lemma \ref{lemmajin}, the last line in \eqref{equ:trace} equals to
    \begin{equation}
        \begin{aligned}
        &\lambda^2 \left\|\begin{bmatrix} \sum_{i=0}^{N-k-1} \mulambda{1}^i\mulambda{2}^{N-k-1-i} & -\mulambda{1}\mulambda{2}\sum_{i=0}^{N-k-2} \mulambda{1}^i\mulambda{2} ^{N-k-2-i}  \end{bmatrix}\right\|^2 \\
        &+\lambda^2\left\|\begin{bmatrix}\sum_{i=0}^{N-k-2} \mulambda{1}^i\mulambda{2}^{N-k-2-i} & -\mulambda{1}\mulambda{2}\sum_{i=0}^{N-k-3} \mulambda{1}^i\mulambda{2}^{N-k-3-i}\end{bmatrix}\right\|^2.
        \end{aligned}
    \end{equation}
    Define $\alambda=|\mulambda{1}| = \sqrt{(1-\beta)(1-h\lambda)}$. By the choice of $\beta$, we have $a_\lambda \le 1-\frac{\sqrt{h\mu}}{2}$. We have the following equation:
    \begin{equation}
        \begin{aligned}
            \lambda^2 \left\|\begin{bmatrix} \sum_{i=0}^{N-k} \mulambda{1}^i\mulambda{2}^{N-k-i} & -\mulambda{1}\mulambda{2}\sum_{i=0}^{N-k-1} \mulambda{1}^i\mulambda{2} ^{N-k-1-i}  \end{bmatrix}\right\|^2 \le 4\lambda^2(N-k)^2 \alambda^{N-k}.
        \end{aligned}
    \end{equation}
    From the definition of $\y^k$ and Cauchy-Schwartz inequality, we have 
    \begin{equation}
        \|\y^k\|_{\A^2}^2\le 8\|\x^k\|_{\A^2}^2+2\|\x^{k-1}\|_{\A^2}^2\le8\|\z^k\|_\C^2+2\|\z^{k-1}\|_\C^2.
    \end{equation}

    Therefore, 
    \begin{equation}
        \begin{aligned}
            &\quad\E \left\|\sum_{k=0}^{N-1} \B^{N-k-1} \beps^k\right\|_\C^2 \\
            &\le 3\sum_{k=0}^{N-1}\sum_{i=1}^{d} 8\lambda_i^2(N-k)^2a_{\lambda_i}^{N-k}\cdot \frac{\|\y^k\|_{\A^2}^2}{m}\\
            &= 24\sum_{i=1}^d \sum_{k=0}^{N-1} \lambda_i^2(N-k)^2a_{\lambda_i}^{N-k}\cdot \frac{\|\y^k\|_{\A^2}^2}{m}\\
        \end{aligned}
        \label{equ:sumexponential}
    \end{equation}

    Then we calculate $\|\B^N \z^0\|_\C^2$. As $\x_{-1}=\x_0$, the contribution of an eigen-directions of $\A$ to the norm is 
    \begin{equation}
        \begin{aligned}
            \lambda^2 x_\lambda^2\left\|\B_\lambda^N \begin{bmatrix}
                1\\1
            \end{bmatrix} \right\|^2,
        \end{aligned}
    \end{equation}
    where $\lambda$ is the eigenvalue, and $x_\lambda$ is the coefficient of the eigen-decomposition of $\x_0$. By Lemma \ref{lemmajin}, we have
    \begin{equation}
        \begin{aligned}
            \B_\lambda^N \begin{bmatrix}
                1\\1
            \end{bmatrix}
            &= \begin{bmatrix}
                \sum_{i=0}^N \mulambda{1}^i\mulambda{2}^{N-i} - \mulambda{1}\mulambda{2} \sum_{i=0}^{N-1} \mulambda{1}^i\mulambda{2}^{N-1-i}\\
                \sum_{i=0}^{N-1}\mulambda{1}^i\mulambda{2}^{N-1-i} - \mulambda{1}\mulambda{2} \sum_{i=0}^{N-2} \mulambda{1}^i\mulambda{2}^{N-2-i}
            \end{bmatrix}\\
            &= \frac{1}{2}\begin{bmatrix}
                \mulambda{1}^N+\mulambda{2}^N+(2-\mulambda{1}-\mulambda{2})\sum_{i=0}^N \mulambda{1}^i\mulambda{2}^{N-i}\\
                \mulambda{1}^{N-1}+\mulambda{2}^{N-1}+(2-\mulambda{1}-\mulambda{2})\sum_{i=0}^{N-1} \mulambda{1}^i\mulambda{2}^{N-1-i}
            \end{bmatrix}\\
            &=\frac{1}{2}\begin{bmatrix}
                \mulambda{1}^N+\mulambda{2}^N+(2-\mulambda{1}-\mulambda{2})\frac{\mulambda{1}^{N+1}-\mulambda{2}^{N+1}}{\mulambda{1}-\mulambda{2}}\\ &\\
                \mulambda{1}^{N-1}+\mulambda{2}^{N-1}+(2-\mulambda{1}-\mulambda{2})\frac{\mulambda{1}^{N}-\mulambda{2}^{N}}{\mulambda{1}-\mulambda{2}}
            \end{bmatrix}\\
            \label{equ:errorterm}
        \end{aligned}
    \end{equation}
    The $\frac{2-\mulambda{1}-\mulambda{2}}{\mulambda{1}-\mulambda{2}}$ term in \eqref{equ:errorterm} can be bounded as follows:
    \begin{equation}
        \begin{aligned}
            \frac{2-\mulambda{1}-\mulambda{2}}{\mulambda{1}-\mulambda{2}} &= \frac{2-(2-\beta)(1-h\lambda)}{\sqrt{(1-h\lambda)(h\lambda(2-\beta)^2-\beta^2)}}\\
            &\le \frac{\beta+h\lambda}{\sqrt{\frac14\cdot h\lambda}}\\
            &\le 2+\sqrt{h\lambda}\\
            &\le 3.
        \end{aligned}
    \end{equation}
    Therefore, 
    \begin{equation}
        \begin{aligned}
            \left\|\B_\lambda^N \begin{bmatrix}
                1\\1
            \end{bmatrix} \right\|^2 &\le \left(|\mulambda{1}^{2N}| + |\mulambda{2}^{2N}| + 9|\mulambda{1}^{2N+2}| + 9|\mulambda{2}^{2N+2}| + |\mulambda{1}^{2N-2}| + |\mulambda{2}^{2N-2}| + 9|\mulambda{1}^{2N}| + 9|\mulambda{2}^{2N}| \right)\\
            &\le 40\left(1-\frac{\sqrt{h\mu}}{2}\right)^{2N-2},
        \end{aligned}
    \end{equation}
    and we have 
    \begin{equation}
        \|\B^N\z^0\|_\C^2 \le 40\left(1-\frac{\sqrt{h\mu}}{2} \right)^{2N-2}\|\z^0\|_\C^2.
    \end{equation}

    Finally, we use induction to prove that $\E \|\z^N\|_{\C}^2 < 200(1-b)^N\|\z^0\|_\C^2$ where $b = 1-\frac{\sqrt{h\mu}}{4}$. Suppose that for $k<N$, we have $\E \|\z^k\|_{\C}^2 <200(1-b)^N \|\z_0\|_\C^2$. By \eqref{equ:znexpansion}, we have
    \begin{equation}
        \begin{aligned}
            \E\|\z^0\|_\C^2 &\le 80\left(1-\frac{\sqrt{h\mu}}{2}\right)^{2N-2}\|\z^N\|_\C^2  + 48h^2\sum_{i=1}^d \sum_{k=0}^{N-1} \lambda_i^2(N-k)^2 a_{\lambda_i}^{N-k}\cdot \frac{\|\y^k\|_{\A^2}^2}{m}.
        \end{aligned}
    \end{equation}
    By the definition of $\y^k$ and the assumption for induction, we have 
    \begin{equation}
        \E{\|\y^k\|_{\A^2}^2} \le 2000(1-b)^{N-1}\|\z^0\|_\C^2.
    \end{equation}
    Using the summation result:
    \begin{equation}
        \sum_{k=1}^n k^2 a^k <\frac{1}{(1-a)^3},
    \end{equation}
    we have
    \begin{equation}
    \begin{aligned}
        \E\|\z^N\|_\C^2  &\le 80\left(1-\frac{\sqrt{h\mu}}{2}\right)^{2N-2}\|\z^0\|_\C^2 + 96000(1-b)^{N-1}\sum_{i=1}^d \frac{h^2\lambda_i^2}{\left(1-\frac{a_{\lambda_i}}{b}\right)^3}\frac{\|\z^0\|_\C^2}{m}\\
        &\le 
        80\left(1-\frac{\sqrt{h\mu}}{2}\right)^{2N-2}\|\z^0\|_\C^2 +384000(1-b)^{N-1}\sum_{i=1}^d \sqrt{h\lambda_i}\frac{\|\z^0\|_\C^2}{m}.
    \end{aligned}
    \end{equation}
    By $h=\frac{m^2}{14400^2(\sum_i\lambda_i^{1/2})^2}$, we have 
    \begin{equation}
        \begin{aligned}
            \E\|\z^N\|_\C^2  \le 80\left(1-\frac{\sqrt{h\mu}}{2}\right)^{N-1}\|\z^0\|_\C^2 + 40(1-b)^{N-1}\|\z^0\|_\C^2.
        \end{aligned}
    \end{equation}
    Therefore, by $h\mu \le 14400^{-2}$ and induction, we have $\E \|\z^N\|_{\C}^2 < 200(1-b)^{N}\|\z_0\|_\C^2$ holds for positive integers $N$.

    Finally, we have
    \begin{equation}
    \begin{split}
        \|\z^N\|_\C^2  &= (\x^N)^\top \A^2\x^N +(\x^{N-1})^\top \A^2\x^{N-1}\\ 
        &\ge \mu\left((\x^N)^\top \A\x^N + (\x^{N-1})^\top \A\x^{N-1} \right)\\
        &= 2\mu\left(f(\x^N) + f(\x^{N-1})\right),
        \end{split}
    \end{equation}
    and 
    \begin{equation}
    \begin{split}
        \|\z^0\|_\C^2  &= 2(\x^0)^\top \A^2\x^0\\ 
        &\le 2L(\x^0)^\top \A\x^0\\
        &= 4Lf(\x^0).
        \end{split}
    \end{equation}
    Therefore,
    \begin{equation}
    \begin{split}
        \E f(\x^N) &\le \frac{1}{2\mu}\cdot 200(1-b)^N \cdot4Lf(\x^0)\\
        &= 400\cdot\frac{L}{\mu}\cdot\left(1-\frac{m\mu}{57600\sum_i \lambda_i^{1/2}} \right)^N\cdot f(\x^0).
    \end{split}
    \end{equation}
    Thus, we finish our proof of Theorem \ref{thm:CORE-AGD}.
\end{proof}

\section{Differenial Privacy} \label{differentialprivacy}
\subsection{Introduction of Differential Privacy}
In distributed machine learning, privacy has attracted increasing attention. In general, people tend to think about whether the machines will reveal information to attackers. However, in this section we study that when information transmitted (for example, $p_{ij}$ in Algorithm \ref{alg:CORE-GD-minibatch}) is leaked, attackers still has no access to the actual gradient information. Moreover, the privacy argument proposed by our paper is based on the differential privacy \cite{dwork2006differential}.
Usually, there is a trade-off between privacy and accuracy. Since random projection is a differential-private operation, our CORE-GD can naturally satisfy certain differential privacy conditions. Below, we introduce basic definitions and our main result in differential privacy.

First we introduce the definition of adjacent vectors and $(\epsilon,\delta)$-differential privacy as below.

\begin{definition}
    For two vectors $\x$ and $\y$, we say $\x$ and $\y$ are adjacent if they satisfy 
    \begin{equation}
        \Vert \x-\y\Vert \leq \Delta_1\|\x\|.
    \end{equation}
\end{definition}
\begin{definition}
    Given $\epsilon ,\delta \geq 0$, letting the output of an algorithm $M$ with input $\x$ be $M(\x)$, the algorithm $M$ satisfies the $(\epsilon,\delta)-$differential privacy property if for an distinguishable set of outputs $S$, and each adjacent variances pairs $\x$ and $\y$, it holds that 
    \begin{equation}
        \mathcal{P}\left(M(\x) \in S\right) \leq {\exp(\epsilon)} \mathcal{P}(M({\y}) \in S)+\delta .
    \end{equation}
\end{definition}

Intuitively, the differential privacy of an algorithm ensures that if two data are adjacent, with a high probability, one cannot distinguish them from the outputs of the algorithm. We notice that in CORE-based algorithm, if two gradient vectors are not far from each other, then after a random projection, the results will be undistinguished with a high probability. So our algorithm naturally has a certain privacy guarantee. Specificaly, we have the result below.

\begin{theorem}
Under the assumptions and settings in Corollary \ref{cor:linear-model}, 
assume that $\Delta_1 < 0.1$. For any $(\epsilon, \delta)$ satisfying $\epsilon = 20\Delta_1\ln{\frac{1}{\delta}}$, Algorithm \ref{alg:CORE-GD-minibatch} with the released information $p_{ij}$ satisfies $(\epsilon,\delta)-$differential privacy.
\label{thm:differential-privacy}
\end{theorem}

Theorem \ref{thm:differential-privacy} is based on the observation mentioned above. Surprisingly, Theorem \ref{thm:differential-privacy} does not depend on the choice of $m$. We think this is because the random projection is rotational invariant, so the attacker can only learn about the norm of the gradient and have no idea about its direction.


\subsection{Proof of Theorem \ref{thm:differential-privacy}}

For the convenience of our proofs, we first present some properties of $(\epsilon,\delta)$-differential privacy.

\begin{definition}
    For two adjacent variances pairs $\x$ and $\y$, an algorithm $M$ and outputs $o$, the privacy loss is defined as
\begin{equation}
    \mathcal{L} = \ln{\frac{\mathcal{P}(M(\x) = o)}{\mathcal{P}(M(\y) = o)}}.
\end{equation}
\label{definitionloss}
\end{definition} 

\begin{lemma}
    $M$ satisfies $(\epsilon,\delta)-$differential privacy if $\mathcal{P}(\mathcal{L}  >  \epsilon) \leq \delta$.
    \label{lem:privacyloss}
\end{lemma}
\begin{proof}
    Letting $B = \{o : \mathcal{L}  >  \epsilon\}$, we have
    \begin{equation}
    \begin{split}
    \mathcal{P}(M(\x) \in S) &= \mathcal{P}(M(\x) \in S \cap B) +\mathcal{P}(M(\x) \in S - B)\\
    &\overset{a}\leq \mathcal{P}(M(\x) \in B) +\mathcal{P}(M(\x) \in S - B)\\
    &\overset{b} \leq \mathcal{P}(M(\x) \in B) +\exp(\epsilon)\mathcal{P}(M(\y) \in S - B)\\
    &\overset{c} \leq \mathcal{P}(M(\x) \in B) +\exp(\epsilon)\mathcal{P}(M(\y) \in S)\\
    &\overset{d} \leq \exp(\epsilon)\mathcal{P}(M(\y) \in S) + \delta .
    \end{split}
    \end{equation}
\end{proof}
In $\overset{a}\leq$ and $\overset{c}\leq$, we use the fact that $\mathcal{P}(X \in A_1) \leq \mathcal{P}(X \in A_2)$ if $A_1 \subseteq A_2$. In $\overset{b}\leq$, we use Definition \ref{definitionloss}. In $\overset{d}\leq$, we use $\mathcal{P}(\mathcal{L} > \epsilon) \leq \delta$.
And by a similar analysis of Lemma \ref{lem:privacyloss}, we have
\begin{lemma}
    $M$ satisfies $(\epsilon,\delta)-$differential privacy if $\mathcal{P}(\mathcal{L}  <  -\epsilon) \leq \delta$.
    \label{lem:privacyloss2}
\end{lemma}

Now we analyze the differential privacy of CORE compression. If we use CORE to compress an vector $\ba$, we project it to $m$ Gaussian vectors $\gauss_1,\cdots,\gauss_m$, and send the inner products $p_i = \langle\ba, \gauss_i \rangle$. We define 
\begin{equation}
    \bXi = \begin{bmatrix}\gauss_1\cdots\gauss_m\end{bmatrix}^\top,
\end{equation}
and
\begin{equation}
    \p = \begin{bmatrix}
        p_1\cdots p_m
    \end{bmatrix}^\top\in \R^m.
\end{equation}
Therefore, we have $\p=\bXi\ba$. We define the compression as 
\begin{equation}
\begin{aligned}
    C:\R^d&\to\R^m\\
    \ba&\mapsto \p. 
\end{aligned}
\end{equation}
Now we study the distribution of $C(\ba)$ for further analysis. 
\begin{lemma}
 $C(\ba) \sim N\left(\mathbf 0, \|\ba\|^2\I_m\right)$.
 \label{lem:C}
\end{lemma}
\begin{proof} 
By the definition of $C$ and the properties of standard Gaussian distribution, $C(\ba)$ must follows an mean zero Gaussian distribution. We notice that the covariance of $p_i$ and $p_j$ is 
\begin{equation}
\begin{split}
    \E p_ip_j = \E \ba^\top\gauss_i\gauss_j^\top\ba =\begin{cases}
        0 &i\ne j,\\
        \|\ba\|^2 &i=j.
    \end{cases}
\end{split}
\end{equation}
Therefore, the variance of $C(\ba)$ is 
\begin{equation}
    \E C(\ba)C(\ba)^\top = \|\ba\|^2 \I_m.
\end{equation}
\end{proof}

By Lemma \ref{lem:privacyloss}, \ref{lem:privacyloss2} and \ref{lem:C}, we can start the proof of Theorem \ref{thm:differential-privacy}.

\begin{proof}[Proof of Theorem \ref{thm:differential-privacy}] To simplify the representation, let $\sigma_1 = \Vert \nabla f(\x^k)\Vert_2$ and  $\sigma_2 = \Vert \nabla f^{'} (\x^k)\Vert_2$, where  $\nabla f(\x^k)$ and $\nabla f^{'}(\x^k)$ are adjacent. By Lemma \ref{lem:C}, we have $C(\nabla f(\x^k)) \sim N(0, \sigma_1^2\I_m)$ and $C(\nabla f^{'}(\x^k)) \sim N(0, \sigma_2^2\I_m)$. Based on Definition \ref{definitionloss}, the privacy loss is
\begin{equation}
\begin{split}
    \mathcal{L} &= \ln\left({\frac{\sigma_2^{m}}{\sigma_1^{m}}\exp\left({\frac{\|\p\|^2}{2}\left(\frac{1}{\sigma_2^2}-\frac{1}{\sigma_1^2}\right)}\right)}\right)\\
    & = \frac{\|\p\|^2}{2}\left(\frac{1}{\sigma_2^2}-\frac{1}{\sigma_1^2}\right) +m\ln{\frac{\sigma_2}{\sigma_1}}.
\end{split}
\end{equation}
If $\sigma_1 >\sigma_2$ we compute the probability of event $\{\mathcal L>\epsilon\}$, which is equivalent to
\begin{equation}
    \|\p\|^2  >  \frac{2\left(\epsilon-m\ln{\frac{\sigma_2}{\sigma_1}}\right)}{\frac{1}{\sigma_2^2}-\frac{1}{\sigma_1^2}}.
\end{equation}
And if $\sigma_1<\sigma_2$ we compute the probability of event $\{\mathcal L<-\epsilon\}$, which is equivalent to 
\begin{equation}
    \|\p\|^2>\frac{2\left(\epsilon - m\ln \frac{\sigma_1}{\sigma_2} \right)}{\frac{1}{\sigma_1^2}-\frac{1}{\sigma_2^2}}.
\end{equation}
We define 
\begin{equation}
    t = \frac{2\epsilon}{\left|\frac{1}{\sigma_2^2}-\frac{1}{\sigma_1^2}\right|}, 
\end{equation}

so in both cases, we have $\mathcal{P}(\mathcal L>\epsilon)\le \mathcal{P} (\|\p\|^2>t)$ or $\mathcal{P}(\mathcal L<-\epsilon)\le \mathcal{P} (\|\p\|^2>t)$. Noticing that $\|\p\|^2$ is the sum of square of independent identically Gaussian distribution, so we have
\begin{equation}
    \|\p\|^2 \sim \sigma_1^2\chi_m^2,
\end{equation}
where $\chi_m^2$ is chi-square distribution with the degree of freedom $m$. According to tail bound of chi-square distribution, we have
\begin{equation}
\begin{split}
    &\mathcal{P}(\|\p\|^2 > t) \overset{a}\leq \exp{\left(-\frac{t}{20\sigma_1^2}\right)} \overset{b}\leq \delta.
\end{split}
\end{equation}

In $\overset{a}\leq$, we use the tail bound of chi-square distribution. If $X \sim \chi_n^2$, then
\begin{equation}
    \mathcal{P}(X > t \cdot 2n) \leq \exp{\left(-\frac{t\cdot n}{10} \right)}.
\end{equation}
In $\overset{b}\leq$, we use the definition of $\epsilon$ that $\epsilon = 20\Delta_1\ln{\frac{1}{\delta}}$, We have
\begin{equation}
\begin{split}
    t&= 40\Delta_1\ln\frac{1}{\delta}\cdot\frac{\sigma_1^2\sigma_2^2}{|\sigma_1^2-\sigma_2^2|}\\
    &= 40\Delta_1\ln\frac{1}{\delta}\cdot \sigma_1^2\cdot\frac{1}{|\sigma_1^2/\sigma_2^2-1|}\\
    &\ge40\Delta_1\ln\frac{1}{\delta}\cdot \sigma_1^2\cdot\frac{1}{2\Delta_1}\\
    &= 20\ln\frac{1}{\delta}\cdot \sigma_1^2.
\end{split}
\end{equation}

Therefore, we have proven that 
\begin{equation}
    \mathcal P(\mathcal L>\epsilon) \le \mathcal P(\|\p\|^2>t)\le\delta, \qquad \sigma_1 >\sigma_2,
    \label{equ:privacy1}
\end{equation}
and
\begin{equation}
    \mathcal P(\mathcal L<-\epsilon) \le \mathcal P(\|\p\|^2>t)\le\delta, \qquad \sigma_1 < \sigma_2.
    \label{equ:privacy2}
\end{equation}

Based on Lemma \ref{lem:privacyloss} and \ref{lem:privacyloss2}, we obtain that our algorithm satisfies $(\epsilon,\delta)-$differential privacy. Thus we finish the proof of Theorem \ref{thm:differential-privacy}.

\end{proof}

\section{Experiment Description and Discussions}\label{sec:apex}

We conduct experiments to test the CORE method. We train the ridge and logistic regressions on the two datasets: MNIST and covtype. Further, we also train the ResNet18 on the two datasets: CIFAR10 and CIFAR100 to test the effect of our method on neural networks. In this section, we only compare our method with some basic compression technique, for example, Gradient Quantization \cite{seide20141,alistarh2017qsgd, tang20211, wen2017terngrad} and Gradient Sparsity \cite{Aji_2017,lin2017deep}, to verify that our algorithm works. We also do not use other compensation techniques such as feedback. In the future, we will add more comparison and improvements to get better experimental results.

\textbf{Methods.} We have implemented the following three gradient compression methods to compare their convergence rate and communication complexities.\par
\noindent
$\bullet$ Gradient Quantization \cite{seide20141,alistarh2017qsgd, tang20211,wen2017terngrad} and . This method compresses each dimension of the gradient to several bits instead of a 32-bit floating-point number to transmit with some techniques of error feedback to reduce the quantization errors. This method can compress the gradient up to 32 times.\par 
~\\
$\bullet$ Gradient Sparsity \cite{Aji_2017,lin2017deep}. This method only preserves the dimensions that occupy more than a certain proportion of the norm in the gradient to transmit while accumulating other dimensions to the next step. In this step other dimensions are replaced by 0 to sparse the gradient. This method works better on the models with gradients having  dominant components. \par
~\\
$\bullet$ CORE. Our method projects the gradient by common Gaussian random vectors in order to realize dimension reduction, which could compress the gradient by a certain multiple.\par 
~\\
\textbf{Performance Plot.} We design two kinds of performance plots. One uses the number of "passes" of the dataset as the $x$-axis.  Note this also reflects the number of communication round since in our experiments the batch-size for all algorithms are the same. Another uses the number of bits the model transmits as the $x$-axis. Both use the training objective distance to the minimum as the $y$-axis.
~\\

\subsection{Linear Model}
We use the above three methods on the following two datasets downloaded from the LibSVM website \cite{chang2008libsvm}:\par 
$\bullet$ The MNIST dataset (784 features). One dataset about $1*28*28$ pixel handwritten $0-9$ pictures.\par 
$\bullet$ The covtype dataset (54 features). One dataset about some features of a piece of land and the types of vegetation that grows on it. \par 

We use distributed gradient descent and accelerated gradient descent to optimize the logistic regression and ridge regression on different datasets. Though we do not give convergence analysis for CORE-AGD on logistic regression, we find it works empirically.
Considering experiments on a real distributed system typical set the number of machines up to 16 as \textcite{alistarh2017qsgd} but some simulation experiments often set the number of machines much bigger, for example 100 as \textcite{freund2022convergence}, we set the number of machines $N = 50$ as a compromise. We take the algorithm without any gradient compression as the baseline and select learning rate from $\{10^{-k}: k \in \mathbb{Z}\}$. In most cases the learning rate is not necessary to be very small, but noticing that Gradient Quantization may cause relatively large error in the early stage of training, we set the learning rate of the algorithm using this method smaller to ensure convergence. We also compare the same algorithm with and without momentum.\par 
To make comparison across datasets, we normalize every vector by its Euclidean norm to ensure the Euclidean norm of each vector is $1$. \par

The results on linear models are shown in Figure \ref{fig1} and \ref{fig2}. The results show that our method has lower communication costs while ensuring a nearly same convergence rate (communication round). We notice that the Gradient Quantization has a poor effect with linear models. And compared to the Gradient Sparsity, our method has a significant advantage on communication costs. Another result is that our method works better with momentum. 

\begin{figure*}[t]
	\centering
        \vspace{-0.35cm}
        \subfigtopskip = 2pt
        \subfigcapskip = -5pt
	\subfigure[convergence]{
       \begin{minipage}[t]{0.5\linewidth}
			\centering
			\includegraphics[width=3in]{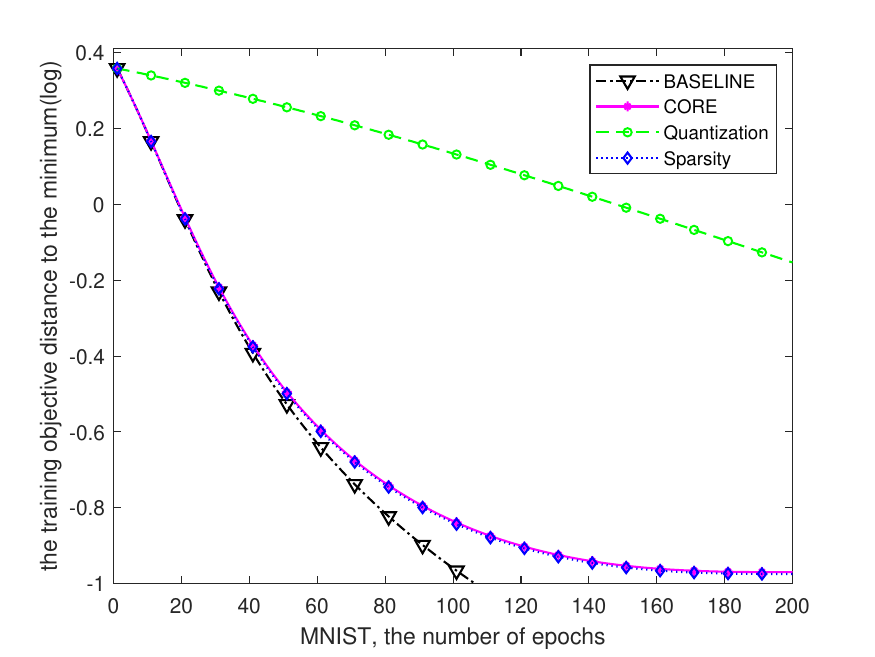}\\			\vspace{0.02cm}
		\end{minipage}%
	}%
	\subfigure[communication]{
		\begin{minipage}[t]{0.5\linewidth}
			\centering
			\includegraphics[width=3in]{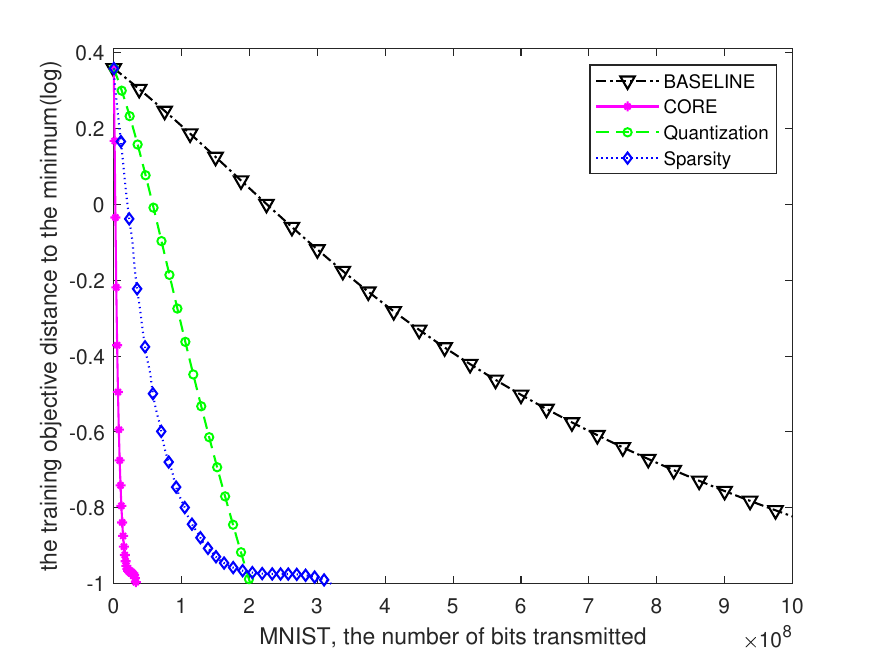}\\
			\vspace{0.02cm}
		\end{minipage}%
	}%
 \\
        \subfigure[convergence]{
       \begin{minipage}[t]{0.5\linewidth}
			\centering
			\includegraphics[width=3in]{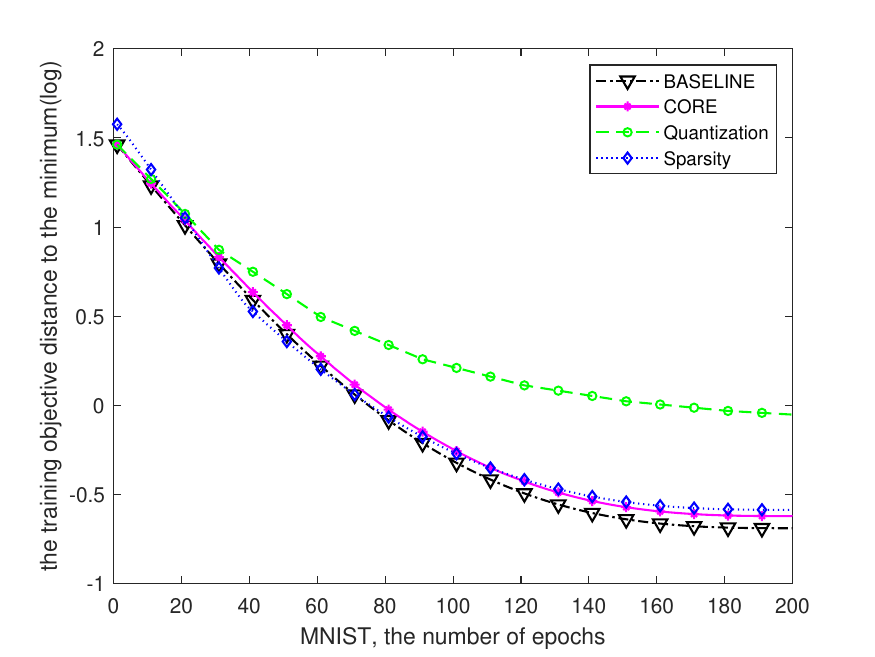}\\
			\vspace{0.02cm}
		\end{minipage}%
	}%
	\subfigure[communication]{
		\begin{minipage}[t]{0.5\linewidth}
			\centering
			\includegraphics[width=3in]{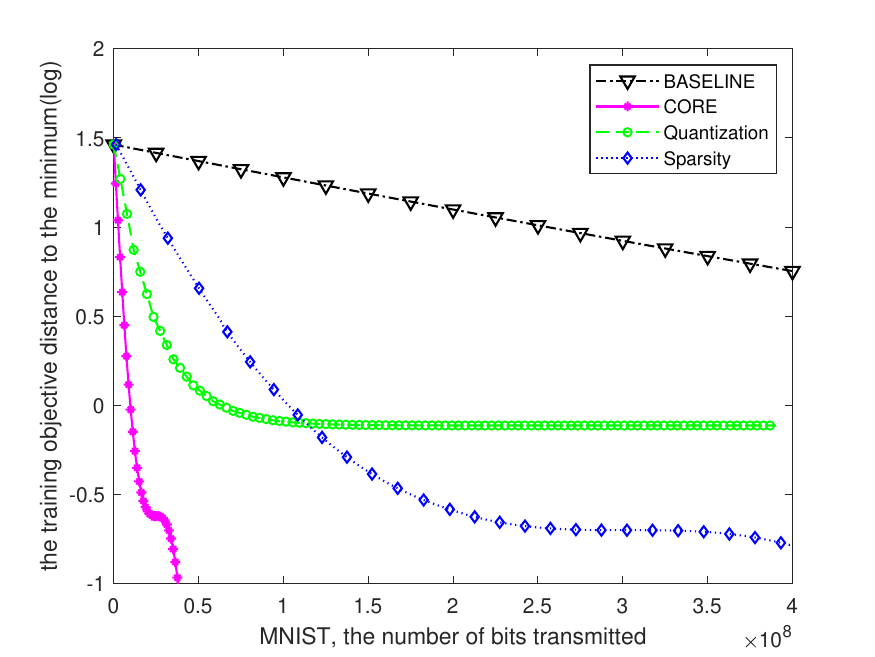}\\
			\vspace{0.02cm}
		\end{minipage}%
	}%
	\centering
	\caption{Experiments on MNIST. (a) and (c) plot the function value against the number of epochs respectively, and (b) and (d) plot the function value against communication costs respectively. (a) and (b) plot the result of logistic regression while (c) and (d) plot the result of ridge regression. }
	\vspace{-0.2cm}
	\label{fig1}
\end{figure*}
\begin{figure*}[t]
	\centering
        \vspace{-0.35cm}
        \subfigtopskip = 5pt
        \subfigcapskip = -5pt
	\subfigure[convergence]{
       \begin{minipage}[t]{0.5\linewidth}
			\centering
			\includegraphics[width=3in]{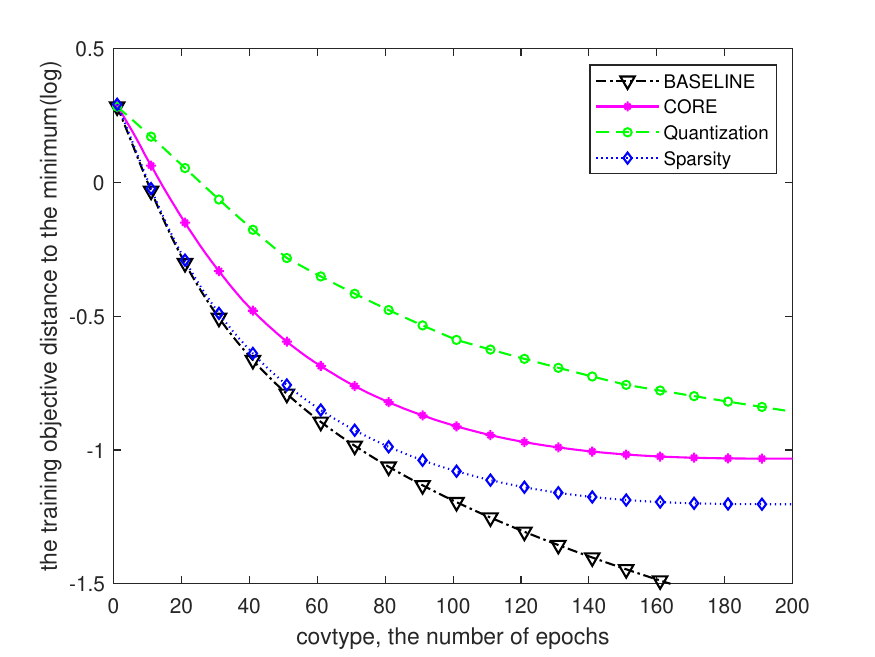}\\			\vspace{0.02cm}
		\end{minipage}%
	}%
	\subfigure[communication]{
		\begin{minipage}[t]{0.5\linewidth}
			\centering
			\includegraphics[width=3in]{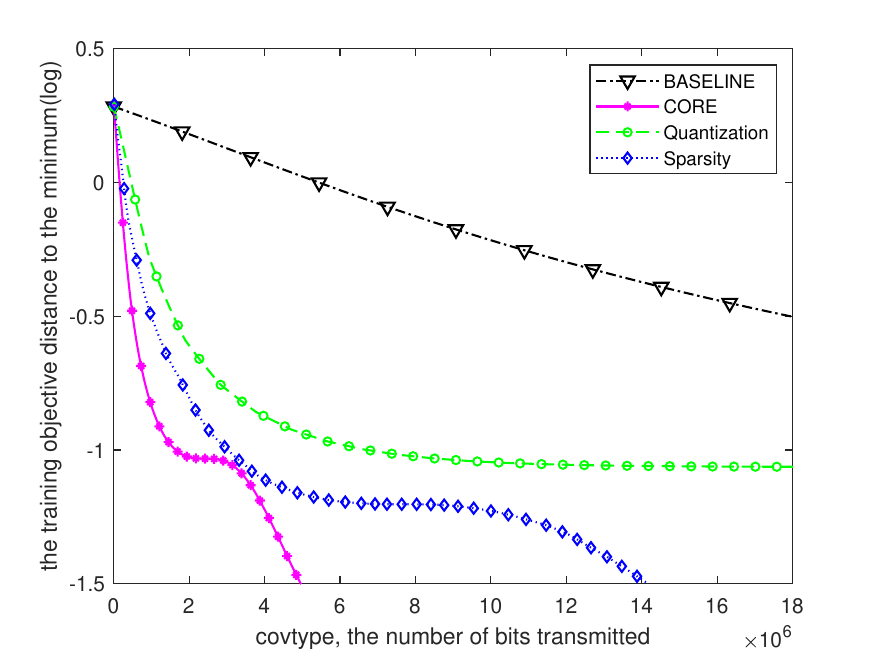}\\
			\vspace{0.02cm}
		\end{minipage}%
	}%
 \\
        \subfigure[convergence]{
       \begin{minipage}[t]{0.5\linewidth}
			\centering
			\includegraphics[width=3in]{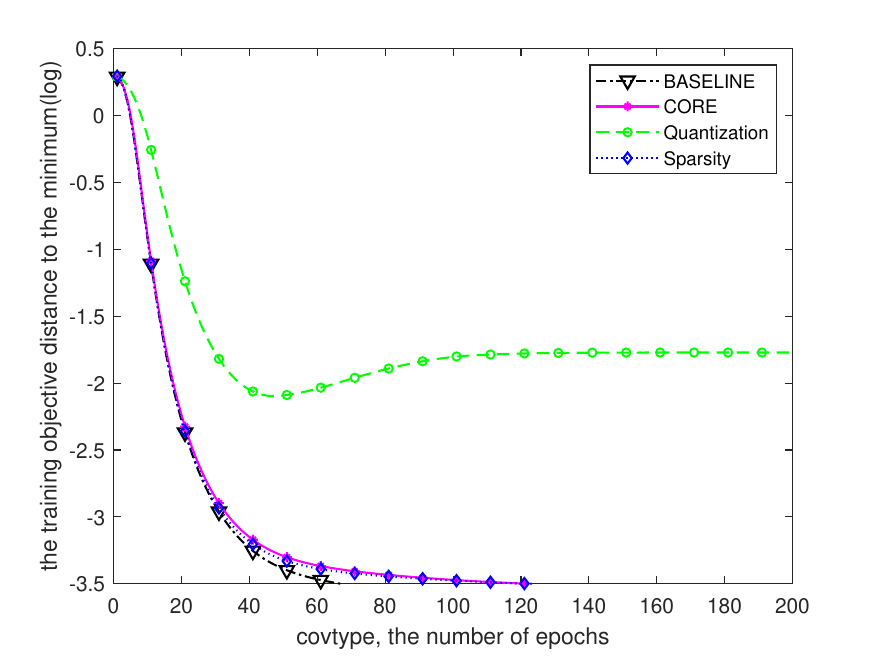}\\
			\vspace{0.02cm}
		\end{minipage}%
	}%
	\subfigure[communication]{
		\begin{minipage}[t]{0.5\linewidth}
			\centering
			\includegraphics[width=3in]{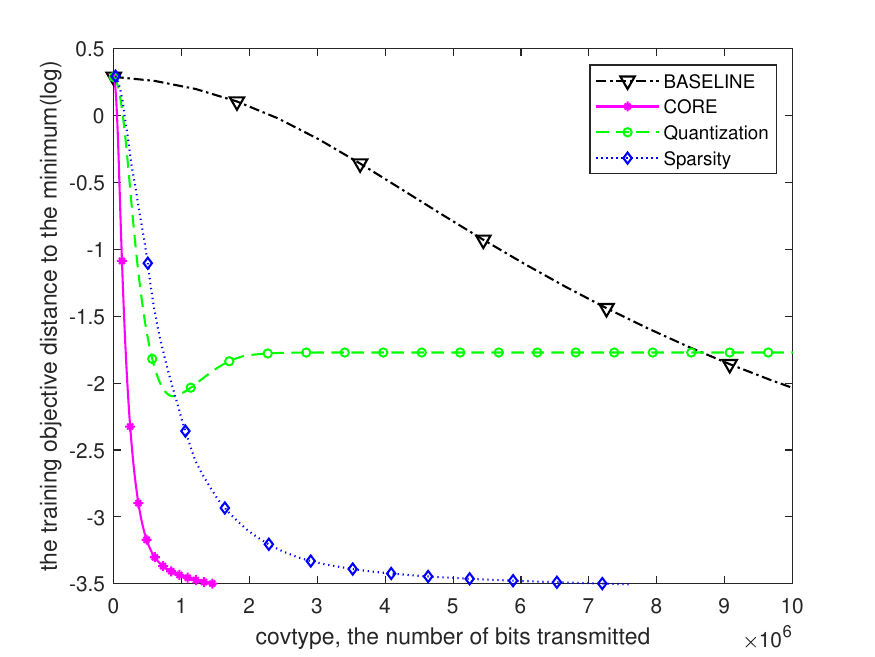}\\
			\vspace{0.02cm}
		\end{minipage}%
	}%
	\centering
	\caption{Experiments on covtype with logistic regression. (a) and (c) plot the function value against the number of epochs without and with momentum, respectively, and (b) and (d) plot the function value against communication costs without and with momentum, respectively.}
	\vspace{-0.2cm}
	\label{fig2}
\end{figure*}

\subsection{Neural Network}
We use the above three methods on the following two datasets downloaded from \url{http://www.cs.toronto.edu/~kriz/cifar.html}:\par 
$\bullet$ the CIFAR10 dataset (50000 samples). One dataset about $3*32*32$ pixel pictures of 10 kinds of different classes.\par 
$\bullet$ the CIFAR100 dataset (50000 samples). One dataset about $3*32*32$ pixel pictures of 100 kinds of classes which can be placed into 20 superclasses.\par 
Our goal is to compare our method with the baseline method, Gradient Quantization and Gradient Sparsity on the speed of convergence and communication costs. Moreover, we also compare CORE with some near results such as PowerSGD \cite{vogels2019powersgd} and DRIVE \cite{vargaftik2021drive}.   
We choose common-used ResNet-18 \cite{he2016deep} as the structure of network. 
We train the model with SGD, the setting of hyperparameters are shown in Table \ref{tab:my_label}. \par

\begin{table}[t]
    \centering
    \begin{tabular}{c|c|c}
         & & \\
         Hyperparameter &CIFAR10 &CIFAR100 \\
         & &\\
         Batch Size(for all machines) &1024 &512 \\
         & &\\
         Batch Size(for each machine) &32 &16 \\
         & &\\
         Machine Numbers &32 &32 \\
         & &\\
         Optimizer &SGD &SGD \\
         & &\\
         Learning Rate &5e-2 &5e-2 \\
         & &\\
         Min Learning Rate &3e-6 &3e-6 \\
         & &\\
         Weight Decay &5e-4 &5e-4\\
         & &\\
         Epoch &200 &200\\
         & &\\
         Learning Rate Scheduler &cosine decay &cosine decay\\
         & &\\
         Input Resolution &32 $\times$ 32 &32 $\times$ 32 \\
         & &\\
         Momentum &0.9 &0.9\\
         & &\\
         Compression Ratio &100+ &80+ \\
         & &
    \end{tabular}
    \caption{Hyperparameter setting of the experiment on networks}
    \label{tab:my_label}
\end{table}
\begin{figure*}[t]
	\centering
        \vspace{-0.35cm}
        \subfigtopskip = 5pt
        \subfigcapskip = -5pt
	\subfigure[convergence]{
       \begin{minipage}[t]{0.5\linewidth}
			\centering
			\includegraphics[width=3in]{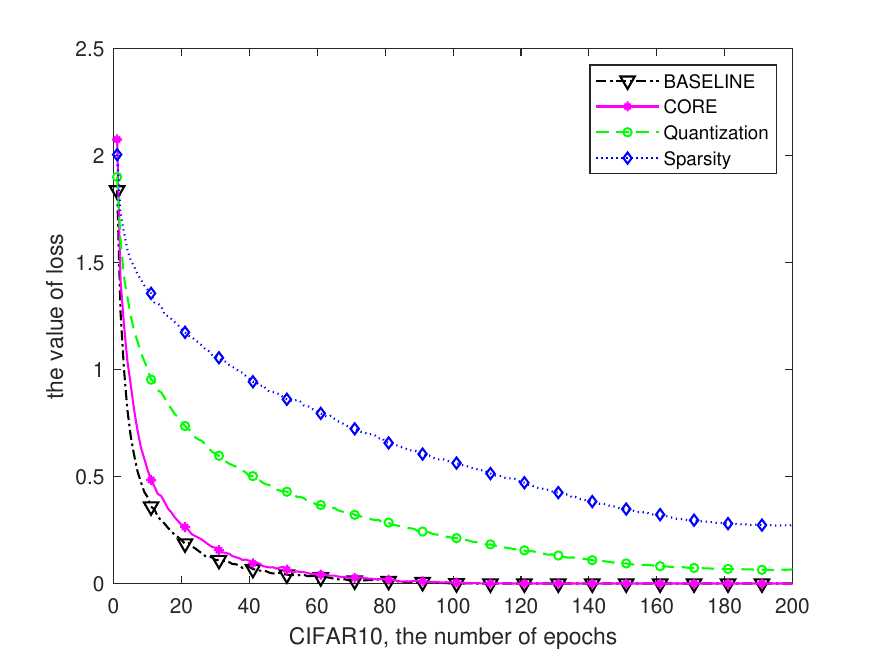}\\
			\vspace{0.02cm}
		\end{minipage}%
	}%
	\subfigure[communication]{
		\begin{minipage}[t]{0.5\linewidth}
			\centering
			\includegraphics[width=3in]{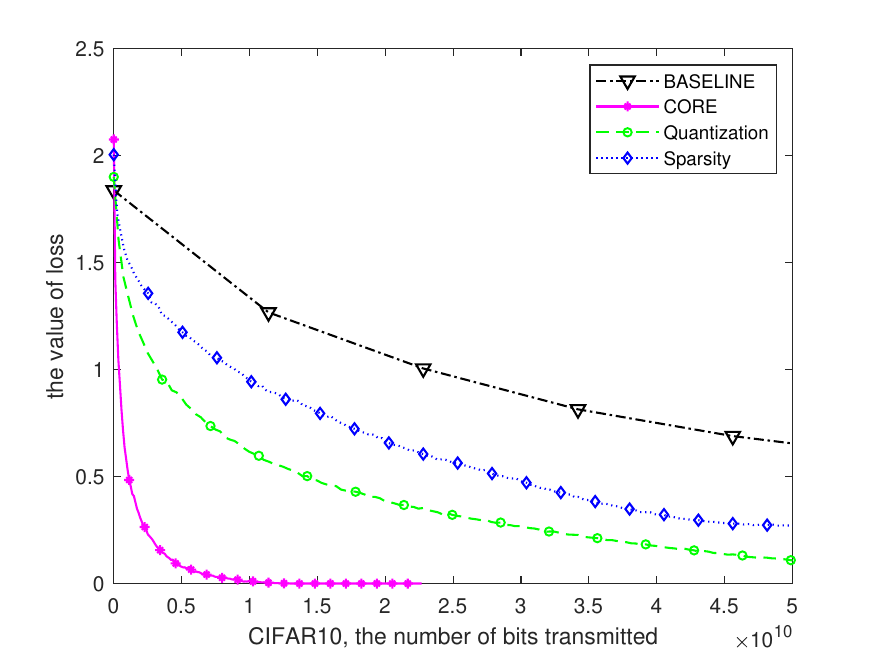}\\
			\vspace{0.02cm}
		\end{minipage}%
	}%
 \\
        \subfigure[convergence]{
        \begin{minipage}[t]{0.5\linewidth}
			\centering
			\includegraphics[width=3in]{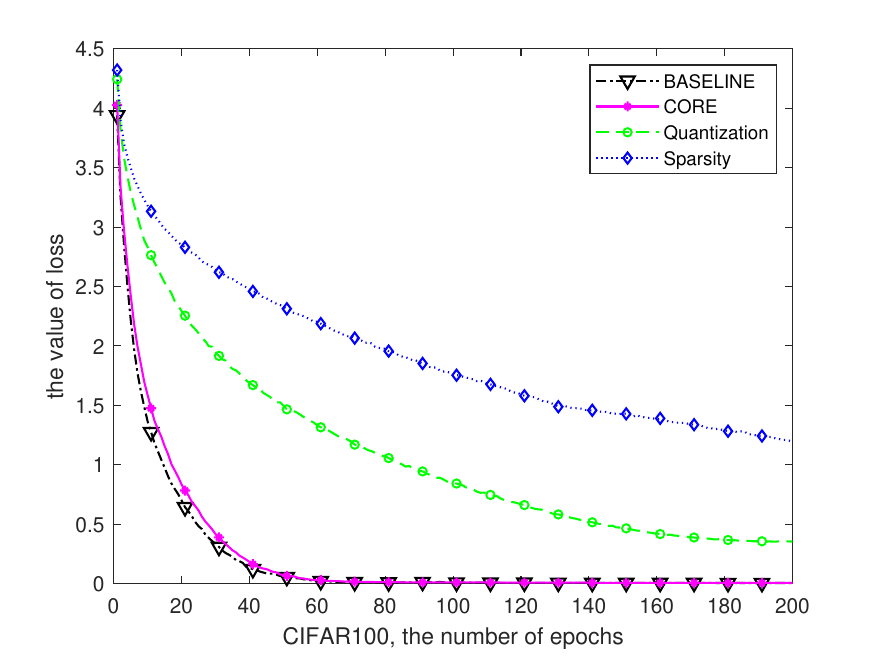}\\
			\vspace{0.02cm}
		\end{minipage}%
	}%
	\subfigure[communication]{
		\begin{minipage}[t]{0.5\linewidth}
			\centering
			\includegraphics[width=3in]{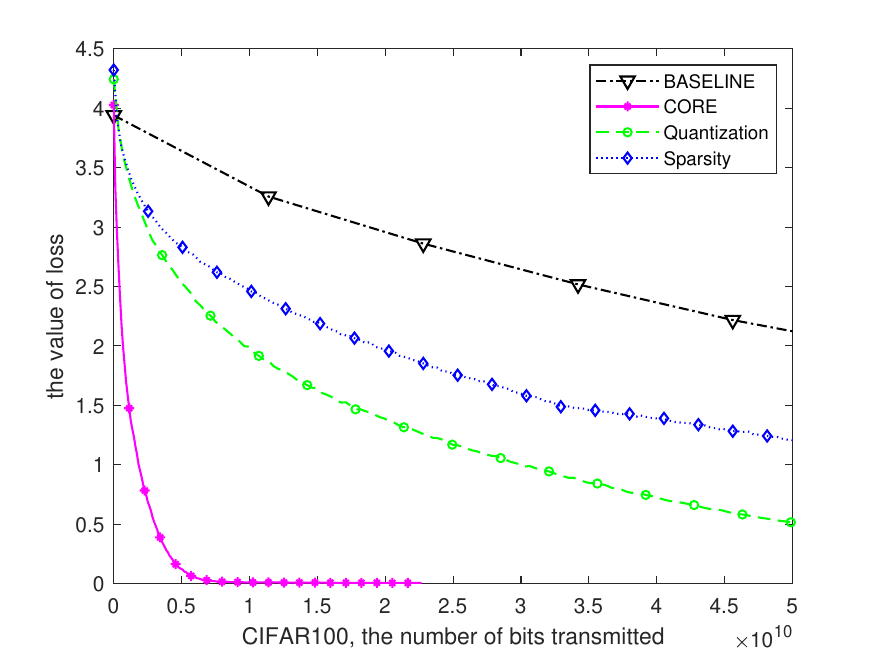}\\
			\vspace{0.02cm}
		\end{minipage}%
	}%
 \\
 \subfigure[convergence]{
        \begin{minipage}[t]{0.5\linewidth}
			\centering
			\includegraphics[width=3in]{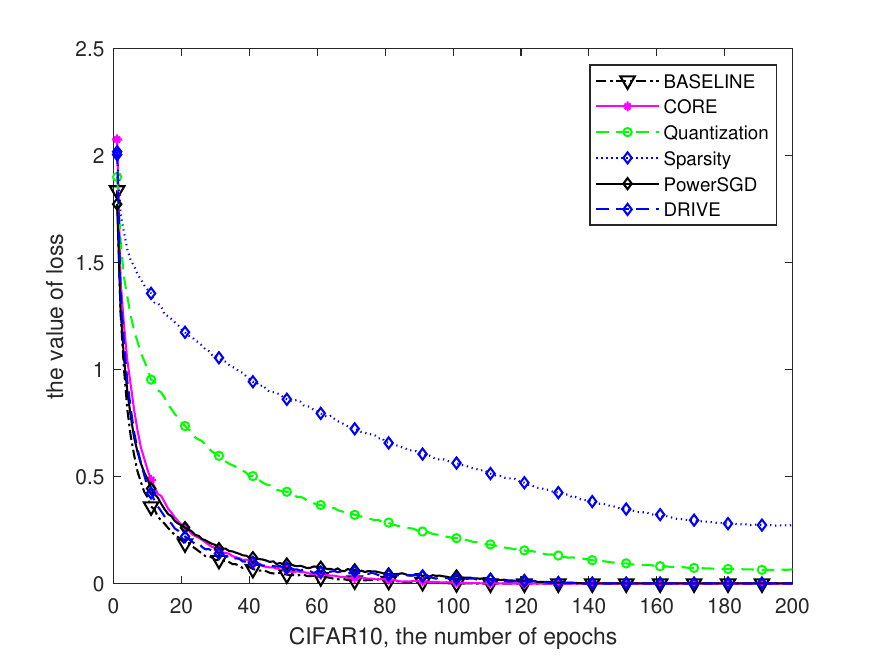}\\
			\vspace{0.02cm}
		\end{minipage}%
	}%
	\subfigure[communication]{
		\begin{minipage}[t]{0.5\linewidth}
			\centering
			\includegraphics[width=3in]{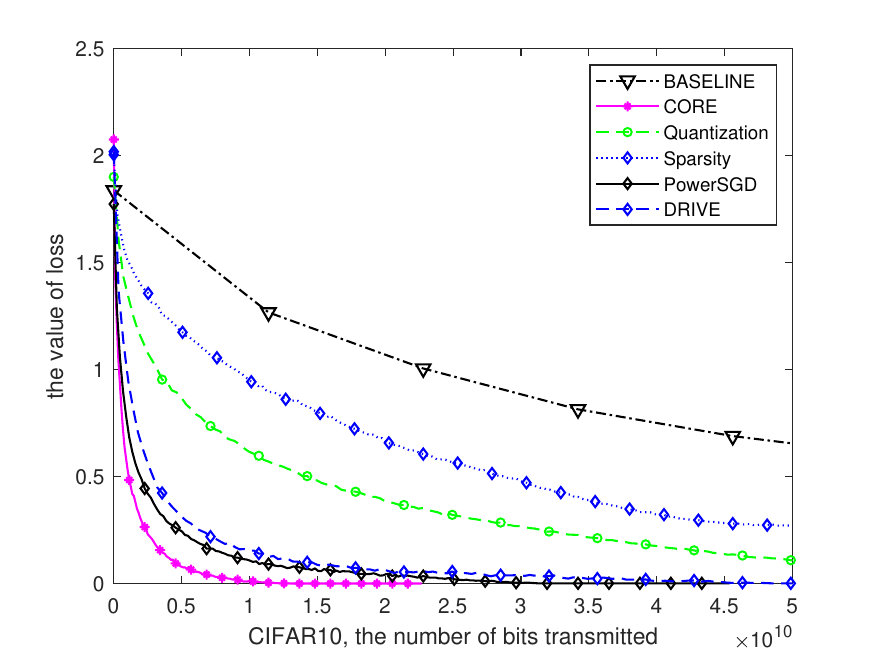}\\
			\vspace{0.02cm}
		\end{minipage}%
	}%
	\centering
	\caption{Experiments on the neural network. (a) and (c) plot the function value against the number of epochs on CIFAR10 and CIFAR100, respectively, and (b) and (d) plot the function value against communication costs on CIFAR10 and CIFAR100, respectively. (e) and (f) present more results compared with PowerSGD \cite{vogels2019powersgd} and DRIVE \cite{vargaftik2021drive}.
 }
	\vspace{-0.2cm}
	\label{fig3}
\end{figure*}

The results on nueral networks are shown in Figure \ref{fig3}. The result shows that our method has a greater convergence rate and communication costs compared to the Gradient Quantization and the Gradient Sparsity. The convergence rate of our method is basically the same as the baseline while the communication costs reduce by hundreds of times. To be more specific, the iteration convergence rate of our CORE method is almost the fastest in the methods participating in the comparison while the number of bits transmitted is much smaller than baseline and almost twice as small as PowerSGD and DRIVE.


\section{Additions}\label{sec: fig}
\subsection{Additional figure}
We show the eigenvalues of data matrix on MNIST and the eigenvalues of a three-layer neural network on MNIST in Figure \ref{fig4}.

\begin{figure*}[t]
	\centering
        \vspace{-0.35cm}
        \subfigtopskip = 10pt
        \subfigcapskip = -5pt
	\subfigure[]{
        \begin{minipage}[t]{0.5\linewidth}
			\centering
			\includegraphics[width=2.5in]{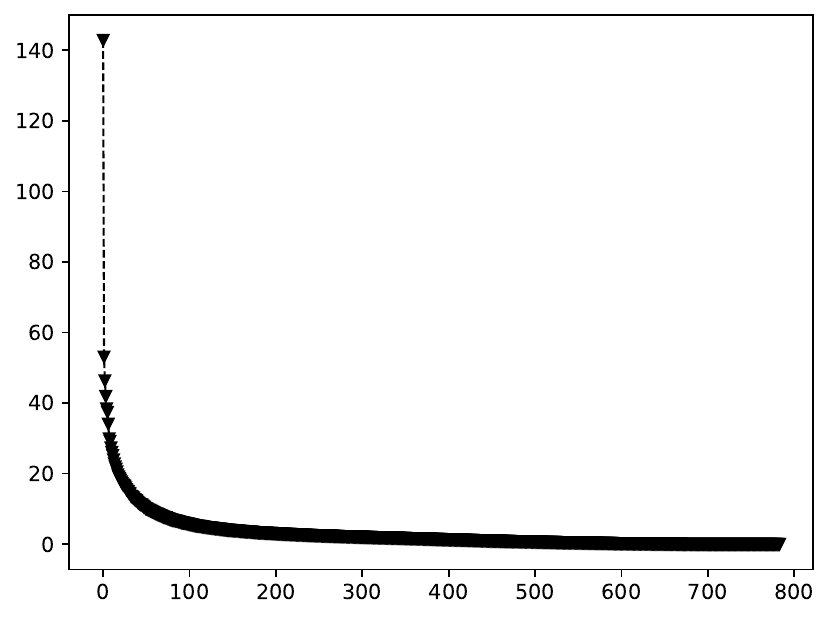}\\			\vspace{0.02cm}
		\end{minipage}%
	}%
	\subfigure[]{
		\begin{minipage}[t]{0.5\linewidth}
			\centering
			\includegraphics[width=2.5in]{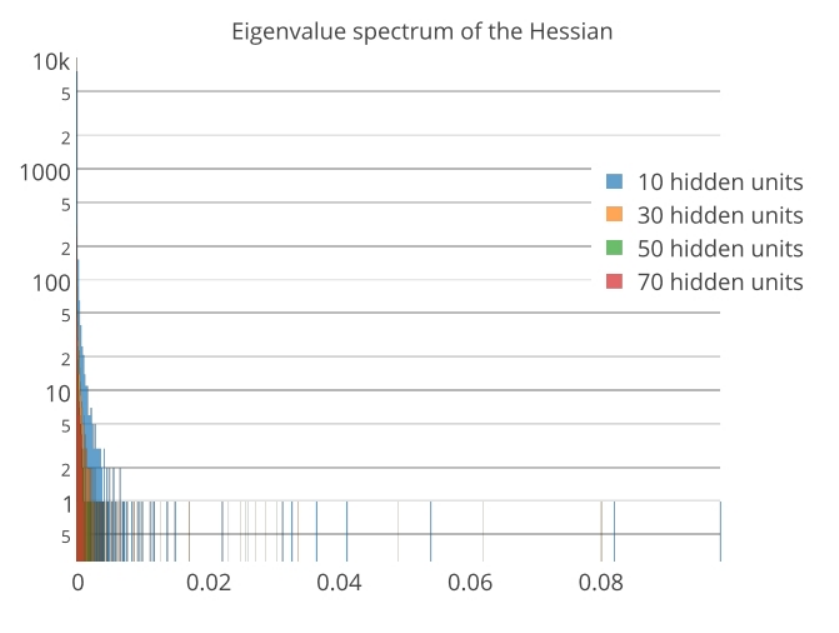}\\
			\vspace{0.02cm}
		\end{minipage}%
	}%
	\centering
	\caption{(a) The eigenvalues of the data matrix on MNIST. (b) The eigenvalues of a three-layer neural network on MNIST. (b) is taken directly from \textcite{sagun2016eigenvalues}. }
	\vspace{-0.2cm}
	\label{fig4}
\end{figure*}

\subsection{More models with dimension-free effective dimension}
We will show more learning models for which the effect dimension is dimension-free.  As one typical example, we consider the two-layer neural network model under suitable conditions.
\begin{proposition}
    Define $f(\W, \w)= \w^\top \sigma(\W^\top\x) $, where $\sigma$ is the activation function. When $\|\x\|_1 \leq r_1$, $\|\w\| \leq r_2$ and $\sigma''(x)\leq \alpha$, we have 
        $\tr\left(\nabla^2 f(\W,\w)\right) \leq \alpha r_1 r_2$.
        \label{prop:2nn}
\end{proposition}

\begin{proof}
    By direct computation, we have
    \begin{equation}
    \begin{aligned}
        \frac{\partial f}{\partial \w} &= \sigma(\W^\top\x),\\
        \frac{\partial f}{\partial \W} &= \left(\sigma'(\W^\top\x)\odot \w\right) \otimes \x,\\
        \frac{\partial^2 f}{\partial \w^2}&= \mathbf 0,\\
        \frac{\partial^2 f}{\partial \W^2} &= \mathop{\mathrm{Diag}}(\sigma''(\W^\top \x)\odot \w)\otimes\x\otimes\x.
    \end{aligned}
    \end{equation}
    Therefore, 
    \begin{equation}
        \begin{aligned}
            \tr\left(\nabla^2 f(\W,\w)\right)) &= \|\x\|^2\cdot \tr\left(\mathop{\mathrm{Diag}}(\sigma''(\W^\top \x)\odot \w)\right)\\
            &\le r_1^2\cdot \langle\sigma''(\W^\top \x), \x \rangle\\
            &\le \alpha r_1r_2.
        \end{aligned}
    \end{equation}
\end{proof}

\end{document}